\documentclass[journal]{IEEEtran}
\usepackage{amsmath,amsfonts}
\usepackage{algorithmic}
\usepackage{algorithm}
\usepackage{array}
\usepackage[caption=false,font=normalsize,labelfont=sf,textfont=sf]{subfig}
\usepackage{textcomp}
\usepackage{stfloats}
\usepackage{url}
\usepackage{verbatim}
\usepackage{graphicx}
\usepackage{cite}
\usepackage{booktabs}
\begin{document}
\bstctlcite{IEEEtranBSTCTL:noDash}

\title{One Stone, Three Birds: Self-adaptive Optimal Transport for Multi-VLM Selection, Adaptation, and Ensembling}

\author{Qiyu Xu, Zhanxuan Hu, Yu Duan, Yonghang Tai, Huafeng Li, Quanxue Gao, and Xiangyong Cao%
\thanks{Corresponding author: Zhanxuan Hu.}
\thanks{Qiyu Xu and Xiangyong Cao are with the School of Computer Science and Technology and the Ministry of Education Key Lab for Intelligent Networks and Network Security, Xi'an Jiaotong University, Xi'an, Shaanxi 710049, China (e-mail: graceafleve@gmail.com; caoxiangyong@xjtu.edu.cn).}
\thanks{Zhanxuan Hu and Yonghang Tai are with the School of Information Science and Technology, Yunnan Normal University, Chenggong, Kunming, Yunnan 650500, China (e-mail: zhanxuanhu@gmail.com; taiyonghang@126.com).}
\thanks{Yu Duan and Quanxue Gao are with the School of Telecommunications Engineering, Xidian University, Xi'an 710071, China (e-mail: duanyue@gmail.com; qxgao@xidian.edu.cn).}
\thanks{Huafeng Li is with Kunming University of Science and Technology, Kunming, China (e-mail: hfchina99@163.com).}}


\maketitle

\begin{abstract}
Vision-language models (VLMs) enable visual recognition from semantic class descriptions, which makes them attractive when target annotations are scarce or unavailable. Most deployment pipelines, however, first choose a single VLM and then adapt that model to the unlabeled target set. This single-backbone paradigm hides a critical assumption: the selected VLM is already compatible with the target domain. In realistic cross-domain deployment, several general-purpose and domain-specialized VLMs may be plausible, yet no instance-level target labels are available to identify the reliable ones. Deployment therefore requires a coupled solution for model selection, target adaptation, and prediction integration.
We revisit this problem from a system-level multi-VLM perspective. Our central observation is that the three decisions above depend on the same latent object: a trustworthy sample-class structure in the target set. Different VLMs may encode different transfer biases and produce conflicting predictions, but their outputs can still provide complementary evidence for estimating this structure.
We propose \textbf{One Stone, Three Birds (OSTB)}, a training-free framework based on self-adaptive optimal transport. Given a pool of frozen candidate VLMs, OSTB estimates a consensus sample-to-class transport plan without updating VLM parameters. The learned transport structure is then reused for all deployment objectives: model selection is performed by ranking the combined semantic and visual reliability induced by the consensus plan; target adaptation is obtained by fitting transport-conditioned visual classifiers; and ensembling is implemented through reliability-aware probabilistic integration.
Extensive experiments on natural-image, remote-sensing, and medical-pathology benchmarks show that OSTB improves model ranking, adaptation stability, and ensemble robustness under heterogeneous candidate pools. Code and data are available at \url{https://github.com/Afleve/OSTB}.

\end{abstract}


\section{Introduction}\label{sec:introduction}
\begin{figure}[t]
	\centering
	\includegraphics[width=\columnwidth]{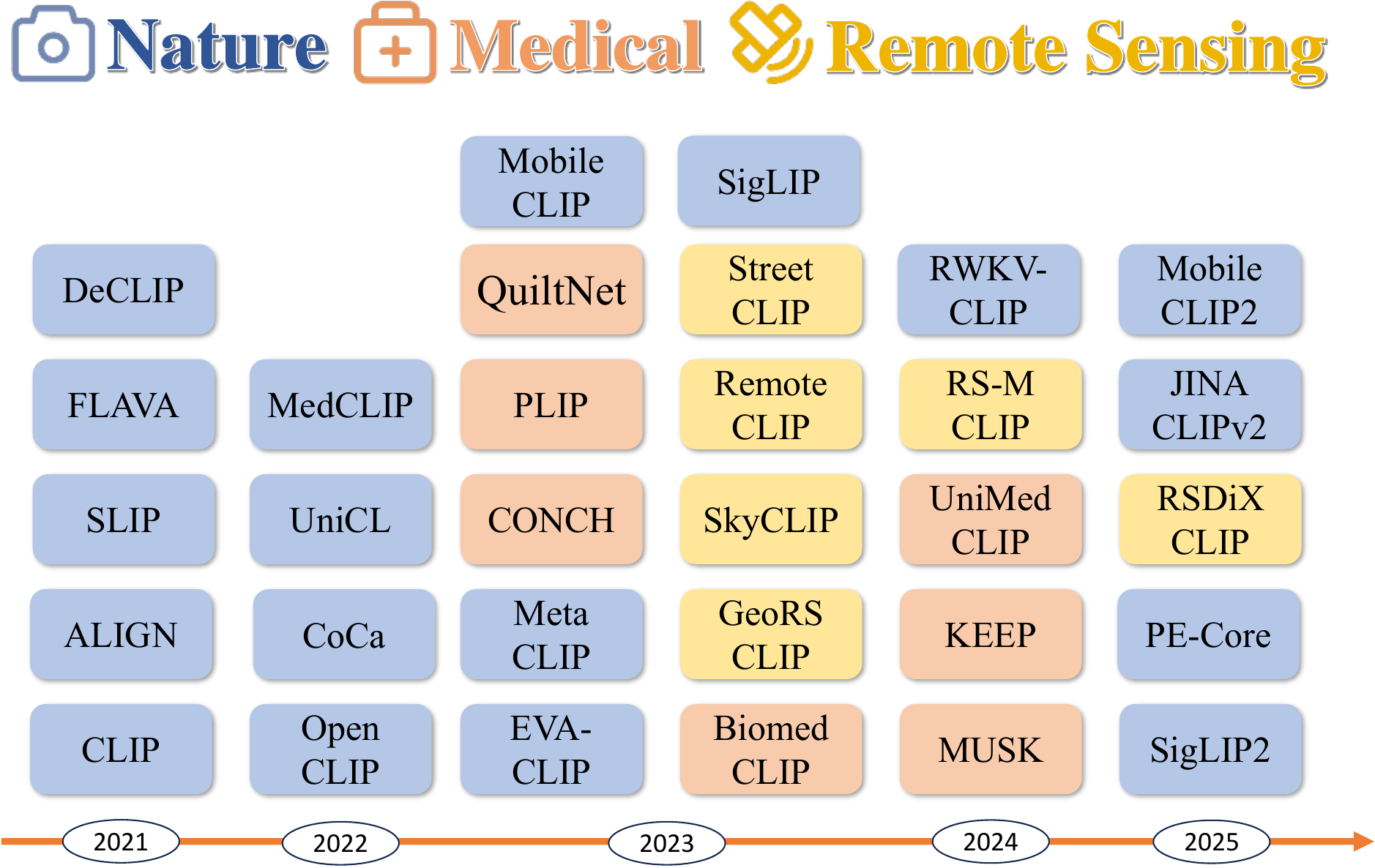}
	\caption{Evolution of the VLM ecosystem across natural-image, remote-sensing, and medical-pathology domains. Diverse pre-training data, objectives, and supervision sources induce heterogeneous transfer biases, making deployment-time model selection nontrivial.}
	\label{Fig:clip_v1_v2}
\end{figure}

\IEEEPARstart{V}{ision-language} models (VLMs) have reshaped visual recognition by connecting images with natural-language class descriptions, enabling recognition tasks to be specified without training a supervised classifier for each label space~\cite{radford2021learning}. This property is particularly useful when target annotations are scarce, since class names may be available even when instance-level labels are not. At the same time, the VLM ecosystem has rapidly diversified, as summarized in Fig.~\ref{Fig:clip_v1_v2}. General-purpose CLIP-style models and domain-oriented variants for remote sensing and medical pathology differ in training data, objectives, and domain priors, which can lead to different transfer behavior across target datasets \cite{siglip,cherti2023reproducible,sun2023evaclip,xu2024demystifying,remoteclip,georsclip,biomedclip,pliphuang2023visual,conchlu2024visual}. Thus, practical deployment is not only about whether VLMs transfer, but also about which available VLMs can be trusted on a particular unlabeled target set.

\begin{figure*}[t]
	\centering
	\includegraphics[width=1\textwidth]{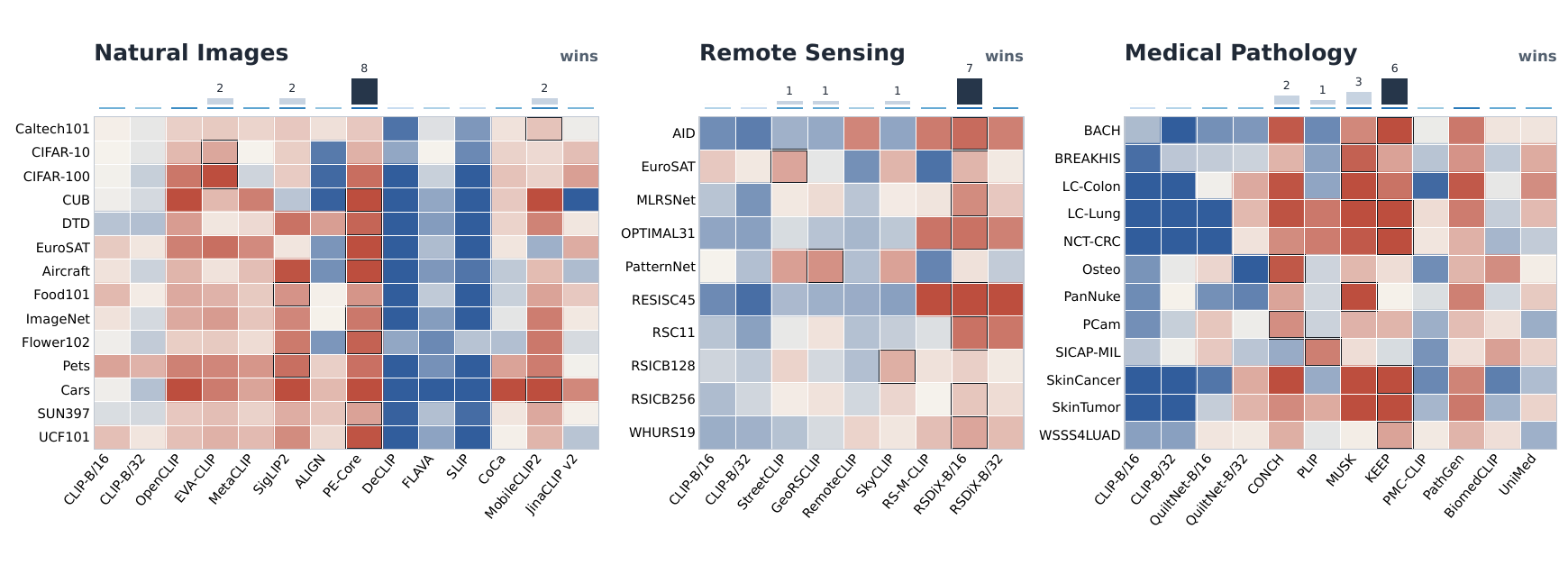}
	\caption{Model-performance fingerprints across natural-image, remote-sensing, and medical-pathology benchmarks. Rows denote datasets and columns denote candidate VLMs. Colors show zero-shot top-1 accuracy relative to the dataset mean, and black boxes mark dataset-wise winners.}
	\label{fig:model_performance_fingerprint}
\end{figure*}

The dominant deployment paradigm still treats VLM adaptation as a single-model problem: a backbone is selected first and then refined by prompt learning~\cite{CoOp,CoCoOp}, classifier adjustment~\cite{lafter}, or test-time and transductive adaptation~\cite{TPT,transductive-3,ZERO,karmanov2024efficient}. This paradigm has become increasingly effective when only unlabeled target data are available, but it largely begins after the candidate model has already been chosen. Once multiple heterogeneous VLMs are available, model choice itself becomes part of deployment. As illustrated in Fig.~\ref{fig:model_performance_fingerprint}, target-dataset compatibility varies substantially across candidate VLMs, and the strongest model can change with visual distribution, label granularity, and semantic composition. Therefore, architecture names, source-domain reputation, and average benchmark performance are insufficient for determining deployment-time reliability without target labels. Moreover, selecting one model, adapting its predictions, and integrating complementary candidates are not independent decisions: an unreliable selected model can corrupt subsequent pseudo-label or structure estimation, single-model adaptation can ignore complementary visual structure encoded by other VLMs, and static ensembling can be dominated by overconfident predictions from mismatched candidates. Multi-candidate VLM deployment therefore raises three coupled questions: \emph{Which models are reliable for the target data? How should their predictions be adapted without labels? How should multiple candidates be integrated when their reliabilities differ?}

Our key insight is that the three decisions can be unified by a shared target-side sample--class structure, which captures both the semantic evidence from candidate VLMs and the visual organization of unlabeled target samples. Optimal transport (OT) offers a natural way to estimate this structure by inducing a soft sample--class coupling \cite{OT}. Compared with simple averaging or local consistency-based assignment, OT can impose both sample and class-prior marginal constraints, discouraging degenerate allocation to a few high-confidence classes while producing a globally consistent sample--class coupling. Existing OT-based methods typically use transport as an alignment or inference mechanism after a model has been selected. In contrast, we use the transport plan to organize the entire multi-candidate deployment process.

To this end, we propose One Stone, Three Birds (OSTB), a training-free framework that jointly performs VLM selection, target adaptation, and model ensembling through a self-adaptive transport plan. OSTB ranks candidate models according to the combined reliability of their semantic posteriors and transport-conditioned visual branches, adapts each candidate using visual classifiers estimated from its image embeddings, and integrates predictions through reliability-aware probabilistic ensembling. Compared with our preliminary conference work SOTA~\cite{hu2026sota}, which focuses on zero-shot fusion across multiple foundation models, this paper studies a broader deployment setting where model ranking, adaptation, and integration must be jointly resolved from the same unlabeled target set. OSTB is model-agnostic and deployment-friendly: it requires no instance-level target annotations, no target validation labels, and no gradient-based fine-tuning of candidate VLMs. Instead, it relies only on the probability outputs and visual representations produced by the available models, making it applicable to heterogeneous, frozen, or black-box-like VLM pools. Experiments on natural-image, remote-sensing, and medical-pathology benchmarks show that OSTB achieves accurate model ranking, effective target-aware adaptation, and robust multi-model prediction. The main contributions are summarized as follows:
\begin{itemize}
\item To the best of our knowledge, this is the first systematic study of deployment-time reliability ranking for cross-domain VLM deployment under a semantic-prior target setting without instance-level target annotations, formulated jointly with target-aware adaptation and reliable ensembling.
\item We propose OSTB, a training-free self-adaptive OT framework in which a shared transport plan serves as the target-side latent structure for candidate ranking, transport-conditioned classifier estimation, and reliability-aware ensemble prediction.
\item We conduct extensive experiments across natural images, remote sensing, and medical pathology, demonstrating strong ranking quality, effective target adaptation, robust ensemble performance, and stable behavior under realistic unlabeled deployment conditions.
\end{itemize}

\section{Related Work}\label{sec:related_work}
\subsection{Vision-Language Models}
CLIP~\cite{radford2021learning} showed that large-scale contrastive image--text pre-training can support open-vocabulary and zero-shot recognition. Later general-purpose VLMs improved this recipe from different angles. ALIGN~\cite{jia2021scaling} scaled weakly aligned image--text data, OpenCLIP~\cite{cherti2023reproducible} made large-scale CLIP-style training reproducible, FLAVA~\cite{singh2022flava} and CoCa~\cite{yu2022coca} introduced broader multimodal pre-training objectives, and EVA-CLIP~\cite{sun2023evaclip}, MetaCLIP~\cite{xu2024demystifying}, JinaCLIPv2~\cite{koukounas2024jinaclipv2}, SigLIP 2~\cite{siglip}, and PE-Core~\cite{bolya2025perception} further improved data curation, optimization, or representation design. As a result, deployment may involve many plausible general-purpose VLMs whose target-domain behavior can differ substantially, even when their architectures or training objectives appear similar.

Domain-specialized VLMs further expand this candidate space. In medical pathology, BiomedCLIP~\cite{biomedclip}, PLIP~\cite{pliphuang2023visual}, CONCH~\cite{conchlu2024visual}, UniMed-CLIP~\cite{khattak2024unimed}, KEEP~\cite{zhou2024keep}, PMC-CLIP~\cite{lin2023pmcclip}, PathGen-CLIP~\cite{sun2025pathgen}, MUSK~\cite{muskxiang2025vision}, and QuiltNet~\cite{ikezogwo2023quilt} adapt vision--language pre-training to pathology imagery and biomedical language. In remote sensing, RemoteCLIP~\cite{remoteclip}, GeoRSCLIP~\cite{georsclip}, SkyCLIP~\cite{skyscript}, RS-M-CLIP~\cite{silva2024multilingual}, RSDiX~\cite{terlizzi2025rsdix}, and StreetCLIP~\cite{haas2023learning} align image--text representations with overhead imagery and geospatial semantics. These models introduce useful domain priors, but also make deployment more ambiguous: before target labels are available, both general-purpose and domain-specialized VLMs may appear plausible. Practical deployment therefore becomes a multi-candidate reliability problem rather than a simple single-backbone choice.

\subsection{The Adaptation of Vision-Language Models}
Most adaptation methods for VLMs assume that the deployment backbone has already been chosen. Early prompt-learning methods such as CoOp~\cite{CoOp} and CoCoOp~\cite{CoCoOp} showed that lightweight text-side adaptation can substantially improve transfer. Later work explored deployment-oriented refinement under unlabeled or test-time settings, including test-time prompt tuning through TPT~\cite{TPT}, label-free classifier refinement through LaFTer~\cite{lafter}, and test-time or transductive adaptation through Transductive Zero-Shot and Few-Shot CLIP~\cite{transductive-3}, ZERO~\cite{ZERO}, Efficient Test-Time Adaptation~\cite{karmanov2024efficient}, COSMIC~\cite{huang2025cosmic}, DPE~\cite{DPE}, and Dual Memory Networks~\cite{zhang2024dual}. These methods improve target-domain performance without standard supervised fine-tuning, but they are still largely designed for the single-model regime.

Among them, COSMIC~\cite{huang2025cosmic} is especially related because it combines the language-aligned semantics of CLIP~\cite{radford2021learning} with the visual structure of DINOv2~\cite{oquab2024dinov2} for robust test-time adaptation. This design shows that target adaptation can benefit from both semantic alignment and auxiliary visual organization. However, COSMIC still adapts a fixed CLIP-style model rather than deciding which VLMs in a heterogeneous pool should be trusted, adapted, or integrated.

Our setting differs in a key respect: deployment may expose several heterogeneous candidate VLMs rather than one fixed backbone. In this case, target adaptation should not only refine predictions, but also help reveal which candidate models are trustworthy for later combination. This makes adaptation, reliability estimation, and model integration coupled deployment decisions rather than independent post-processing steps.

\subsection{The Selection and Integration of Vision-Language Models}
Recent studies have increasingly investigated how multiple foundation models can be composed for downstream visual tasks. In open-vocabulary and promptable segmentation, ODISE~\cite{ODISE}, SEEM~\cite{SEEM}, Grounded-SAM~\cite{GroundedSAM}, and SAM 3~\cite{SAM3} combine VLMs with grounding, segmentation, or concept-level modules so that different pretrained components jointly support broad visual understanding. These methods demonstrate the complementarity of heterogeneous foundation models, but their emphasis is usually on task composition and open-world prediction. They do not directly address how to estimate the reliability of a candidate VLM pool under unlabeled target-domain shift.

Another related line studies model selection through transferability estimation and source-model ensemble selection. Representative criteria include H-score~\cite{HScore}, LEEP~\cite{LEEP}, LogME~\cite{LogME}, and TransRate~\cite{TransRate}. PARC~\cite{Bolya2021PARC} studies scalable diverse model selection, while transferability-based source ensemble construction~\cite{Agostinelli2022,WinningTeam2023} and transferability-metric stability analysis~\cite{Agostinelli2022StableTM} further examine how source models or metrics should be selected. These methods are useful for comparing pre-trained models or constructing source-model subsets, but their outputs usually remain transferability scores, selected subsets, or diagnostic criteria. They do not simultaneously produce a target-adapted classifier and a reliability-aware ensemble predictor for heterogeneous off-the-shelf VLMs under unlabeled target deployment.

Rethinking Model Selection in VLM~\cite{li2026rethinking} uses the Gromov-Wasserstein distance to estimate structural compatibility between candidate vision encoders and a target LLM for VLM construction, showing that model size and zero-shot accuracy can be weak predictors of final VLM performance. This work is closely related in spirit because it also questions simple model-choice heuristics. However, its goal is to select components for VLM construction, whereas OSTB focuses on deployment-time selection among already available frozen VLMs and jointly derives model ranking, target adaptation, and ensembling from unlabeled target data.

Ensemble methods such as deep ensembles~\cite{lakshminarayanan2017simple} and model soups~\cite{wortsman2022model} can exploit model complementarity, but they do not estimate target-side reliability from unlabeled data. Uniform or static combination can therefore be dominated by mismatched models that produce overconfident predictions. OSTB addresses a different deployment setting: candidate VLMs must be ranked, adapted to the target domain, and integrated into a final predictor without instance-level target labels. Rather than combining an existing selector, an existing adapter, and an existing ensemble rule as independent modules, OSTB derives all three from the same target-side structure learned from VLM probability outputs and induced visual representations. This distinction separates our setting from prior model-selection, ensemble-selection, and open-world model-assembly methods.

\subsection{Optimal Transport}
Optimal transport (OT)~\cite{OT} provides a principled way to align distributions while respecting their geometry. It has been used for universal domain adaptation~\cite{chang2022unified}, VLM prompt learning through PLOT~\cite{chenplot}, open-vocabulary recognition through Recover and Match~\cite{tan2025recover}, and test-time transfer through AWT~\cite{awt}. These methods mainly use OT as an alignment or inference module after a model or representation has been specified.

Rather than applying transport only after a backbone has been fixed, OSTB treats the transport plan as a shared latent structure that is jointly informative for model ranking, target-aware adaptation, and multi-model integration. Different from OT formulations that align fixed distributions with predefined costs or weights, OSTB uses agreement with the evolving transport plan to adaptively estimate branch reliability and refine visual GMMs. In this way, OT becomes the organizing mechanism that couples the three deployment goals.

\section{Method}\label{sec:method}
OSTB is organized as a deployment pipeline rather than as a single adaptation rule. Given semantic priors, an unlabeled target adaptation set, and several frozen VLMs, the method estimates which models can be trusted, how the target structure should adapt them, and how their predictions should be combined. We first define the problem setup, then construct semantic and visual posterior views, estimate a shared transport structure, and finally reuse that structure for the three deployment outputs.

\begin{figure*}[t]
    \centering
    \includegraphics[width=0.98\textwidth]{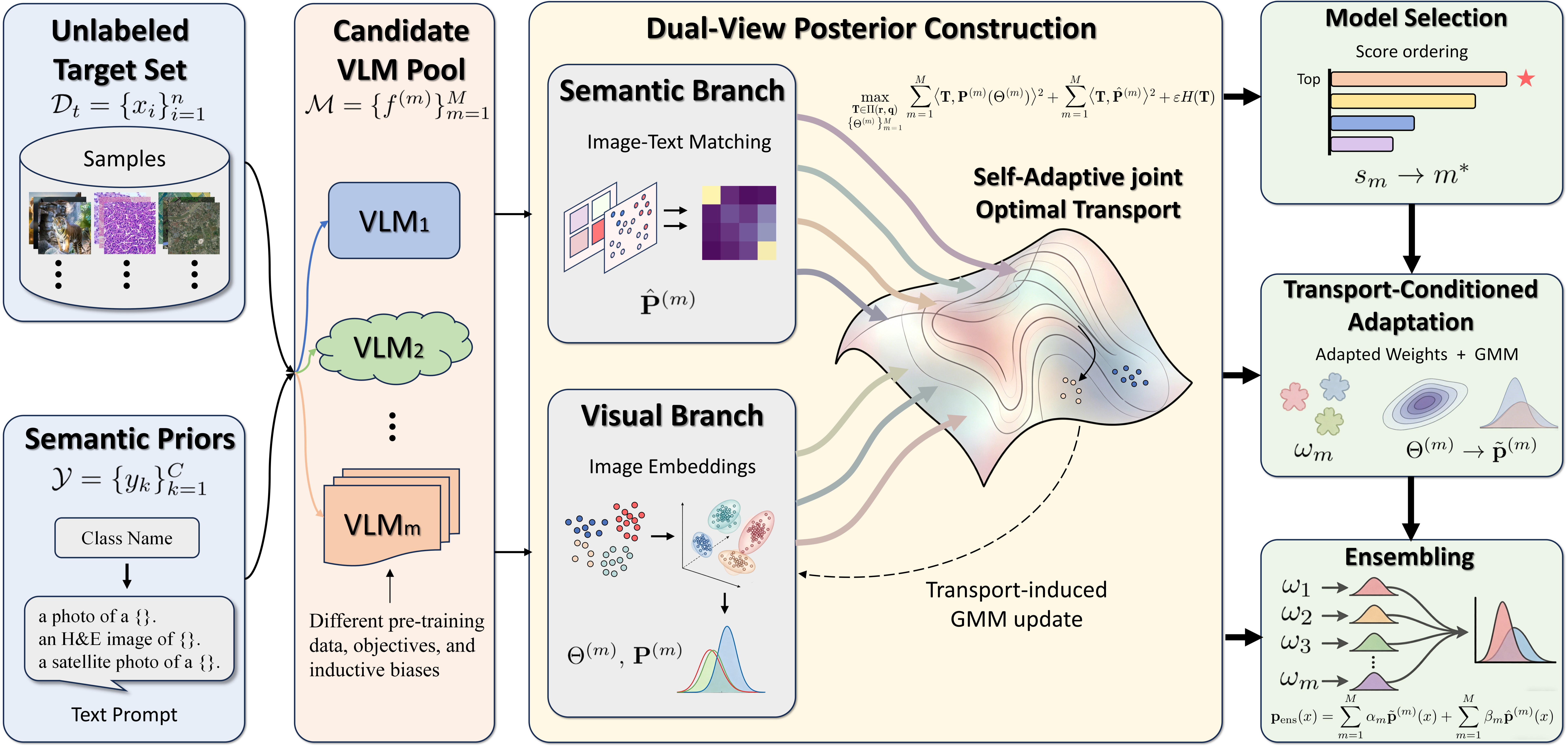}
    \caption{Framework of OSTB. Starting from an unlabeled target adaptation set and semantic priors, OSTB evaluates a heterogeneous pool of candidate VLMs through two complementary views: a semantic branch based on image-text matching and a visual branch based on feature-space GMMs. A self-adaptive optimal transport module integrates the two views, estimates model reliability, and uses the shared target structure to refine the visual GMMs. The resulting information is reused for three deployment goals: selecting reliable models, adapting the candidates to the target domain, and fusing their predictions into a robust ensemble.}
    \label{fig:ostb_framework}
\end{figure*}

\subsection{Problem Setup}
We consider cross-domain deployment under a class-name-only target setting. Let $\mathcal{D}_{t}=\{x_i\}_{i=1}^{n}$ denote the unlabeled target adaptation set, and let $\mathcal{Y}=\{y_k\}_{k=1}^{C}$ denote the target label set, where each class is specified only by its category name. No instance-level target annotations or target validation labels are available during deployment.

We are given a candidate model pool $\mathcal{M}=\{f^{(m)}\}_{m=1}^{M}$, where $M$ is the number of frozen candidate VLMs. These candidates may differ in pre-training data, objective, scale, and domain specificity, and can therefore induce substantially different target-domain behavior. The deployment goal is to derive three outputs from this unlabeled target adaptation set: a reliability ranking over candidate VLMs, target-aware classifiers for adaptation, and a final ensemble predictor. The setup assumes no external validation labels, supervised target fine-tuning, or oracle knowledge of which candidate VLM is best for the target domain.

This setting differs from standard single-model adaptation. Each VLM can already produce class-name-based predictions, so the key question is not how to obtain a prediction, but how to judge whether that prediction is compatible with the target distribution. OSTB therefore treats reliability estimation, target adaptation, and multi-model integration as coupled consequences of one latent target structure rather than as independent post-processing steps.

\subsection{Dual-View Posteriors from Candidate VLMs}
To estimate the latent target structure, OSTB first expresses different sources of evidence in a common sample--class probability form. For each candidate VLM, we construct two complementary posterior views. The semantic view is directly derived from image--text similarity and is therefore naturally aligned with target class names. However, under domain shift, the learned alignment between visual embeddings and textual class prototypes may become unreliable, causing the semantic posterior to overlook useful target-side visual structure. We therefore also construct a visual view from the image features themselves and convert it into label-indexed probabilities.
\paragraph{Semantic posteriors from VLMs}
For the $m$-th VLM, let $g_I^{(m)}$ and $g_T^{(m)}$ denote its image and text encoders. Given image embedding $\mathbf{z}_i^{(m)}=g_I^{(m)}(x_i)$ and class-text embedding $\mathbf{t}_k^{(m)}=g_T^{(m)}(y_k)$, we obtain the semantic posterior by
\begin{equation}
	\hat{p}_{ik}^{(m)}=
	\frac{\exp\left(\tau_m\cdot \mathrm{sim}\left(\mathbf{z}_i^{(m)},\mathbf{t}_k^{(m)}\right)\right)}
	{\sum_{j=1}^{C}\exp\left(\tau_m\cdot \mathrm{sim}\left(\mathbf{z}_i^{(m)},\mathbf{t}_j^{(m)}\right)\right)},
	\label{eq:semantic_prob}
\end{equation}
where $\tau_m$ is the temperature parameter and $\mathrm{sim}(\cdot,\cdot)$ denotes cosine similarity. Stacking all $\hat{p}_{ik}^{(m)}$ yields the row-stochastic matrix $\hat{\mathbf{P}}^{(m)}$. This matrix serves as the semantic anchor of OSTB because its columns are explicitly tied to the target category names. Nevertheless, using it alone still assumes that the selected VLM provides reliable image--text alignment on the target domain, which may not hold under domain shift.

\paragraph{Visual posteriors from VLM image features}
The visual branch complements the semantic posterior by modeling the target-side structure encoded in each VLM image feature space. Although a VLM can output class probabilities through image--text similarity, these probabilities may be biased when the semantic and visual embedding spaces are not well aligned on the target domain. In such cases, the image encoder may still preserve meaningful cluster geometry among unlabeled samples, but this visual structure can be weakened if one relies only on the semantic posterior. Since raw image features are not label-indexed probabilities and cannot be directly fused with Eq.~\eqref{eq:semantic_prob}, we introduce lightweight class-wise Gaussian mixtures to convert the visual structure into a probabilistic form over the same label set.

Concretely, for each model $m$, we reuse the image embedding $\mathbf{z}_i^{(m)}\in\mathbb{R}^{d_m}$ and parameterize a Gaussian mixture
\begin{equation}
	\Theta^{(m)}=\left\{\{\pi_k^{(m)},\boldsymbol{\mu}_k^{(m)}\}_{k=1}^{C},\boldsymbol{\Sigma}^{(m)}\right\},
\end{equation}
where one Gaussian component is associated with each target class. The class association is initialized from the multi-VLM semantic consensus rather than from unconstrained clustering. Specifically, we average the zero-shot probability outputs of the candidate VLMs and use the resulting semantic consensus as soft responsibilities when initializing the class-associated Gaussian statistics in each VLM feature space. This initialization aligns the otherwise label-permutation-invariant GMM components with the shared target label set before they are further refined by transport-induced assignments. The corresponding posterior probability is
\begin{equation}
	p_{ik}^{(m)}\!\left(\Theta^{(m)}\right)=
	\frac{\pi_k^{(m)}\mathcal{N}\!\left(\mathbf{z}_i^{(m)}\mid\boldsymbol{\mu}_k^{(m)},\boldsymbol{\Sigma}^{(m)}\right)}
	{\sum_{j=1}^{C}\pi_j^{(m)}\mathcal{N}\!\left(\mathbf{z}_i^{(m)}\mid\boldsymbol{\mu}_j^{(m)},\boldsymbol{\Sigma}^{(m)}\right)},
	\label{eq:visual_prob}
\end{equation}
where $\mathcal{N}(\cdot)$ denotes the Gaussian density. Stacking these probabilities yields $\mathbf{P}^{(m)}(\Theta^{(m)})\in[0,1]^{n\times C}$. For convenience and efficiency, we use shared diagonal covariance statistics in the visual Gaussian branch.

\subsection{Self-Adaptive Joint Optimal Transport}
After the two posterior views are built, the remaining question is how to form a target consensus without knowing which models or branches are reliable. Direct averaging would assign equal trust to all candidates, while choosing one model would discard useful complementary evidence. OSTB instead estimates a transport plan whose sample-class assignments agree more strongly with reliable posteriors and weakly with unreliable ones. Following optimal transport (OT) \cite{OT}, we define $\mathbf{r}\in\Delta_n$ as the marginal distribution over unlabeled target adaptation samples and $\mathbf{q}\in\Delta_C$ as the marginal distribution over semantic categories, where $r_i$ and $q_k$ denote the marginal probabilities of the $i$-th sample and the $k$-th semantic category, respectively. Here $\mathbf{1}_{n}\in\mathbb{R}^{n}$ and $\mathbf{1}_{C}\in\mathbb{R}^{C}$ denote all-one vectors. Since no labels or external class-frequency estimates are available for the target adaptation set, both marginal distributions are set to be uniform.
The feasible set of transport plans is therefore
\begin{equation}
\Pi(\mathbf{r},\mathbf{q})=
\left\{\mathbf{T}\in\mathbb{R}_{+}^{n\times C} \mid \mathbf{T}\mathbf{1}_{C}=\mathbf{r},\ \mathbf{T}^{\top}\mathbf{1}_{n}=\mathbf{q}\right\}.
\label{eq:transport_set}
\end{equation}

If the reliability of each branch were known in advance, a natural integration strategy would be to solve the weighted entropic OT problem
\begin{equation}
\begin{aligned}
\max_{\mathbf{T}\in\Pi(\mathbf{r},\mathbf{q})}\quad
&\sum_{m=1}^{M}\alpha_m\left\langle \mathbf{T},\mathbf{P}^{(m)}(\Theta^{(m)})\right\rangle \\
&+\sum_{m=1}^{M}\beta_m\left\langle \mathbf{T},\hat{\mathbf{P}}^{(m)}\right\rangle
+\varepsilon H(\mathbf{T}),
\end{aligned}
\label{eq:weighted_ot}
\end{equation}
where $\boldsymbol{\alpha},\boldsymbol{\beta}\in\mathbb{R}_{+}^{M}$ are branch-model reliability weights satisfying $\sum_{m=1}^{M}\alpha_m=1$ and $\sum_{m=1}^{M}\beta_m=1$, $H(\mathbf{T})=-\sum_{i,k}T_{ik}\log T_{ik}$ is the entropy regularizer \cite{OT}, and $\langle \mathbf{A},\mathbf{B}\rangle=\mathrm{tr}(\mathbf{A}^{\top}\mathbf{B})$ denotes the Frobenius inner product.

The difficulty is that these reliabilities are exactly what cannot be determined from labeled target validation data. To remove manual weight tuning, we adopt the self-adaptive objective
\begin{equation}
\max_{\substack{\mathbf{T}\in\Pi(\mathbf{r},\mathbf{q})\\\{\Theta^{(m)}\}_{m=1}^{M}}}
\sum_{m=1}^{M}
\left\langle \mathbf{T},\mathbf{P}^{(m)}(\Theta^{(m)})\right\rangle^{2}
+\sum_{m=1}^{M}
\left\langle \mathbf{T},\hat{\mathbf{P}}^{(m)}\right\rangle^{2}
+\varepsilon H(\mathbf{T}).
\label{eq:self_adaptive_objective}
\end{equation}
Equation~\eqref{eq:self_adaptive_objective} unifies semantic distributions and visual GMM-induced distributions under a single transport plan. The plan is shaped by the label-aligned semantic branch and the visually coherent auxiliary branch, while the visual branch itself is refined by transport-induced assignments. This closed-loop interaction is the key mechanism through which OSTB exploits multiple candidate VLMs without relying on external visual foundation models.

\subsection{Optimization via Minorization-Maximization}
The quadratic agreement terms in Eq.~\eqref{eq:self_adaptive_objective} make the joint objective nonlinear in both $\mathbf{T}$ and $\{\Theta^{(m)}\}_{m=1}^{M}$. We derive a minorization-maximization (MM) update \cite{MM} for the transport subproblem and alternate it with a transport-conditioned moment refinement of the visual GMMs. At iteration $t$, let
\begin{equation}
a_{v,m}^{(t)}=
\left\langle \mathbf{T}^{(t)},\mathbf{P}^{(m)}\!\left(\Theta^{(m,t)}\right)\right\rangle,
\qquad
a_{s,m}^{(t)}=
\left\langle \mathbf{T}^{(t)},\hat{\mathbf{P}}^{(m)}\right\rangle .
\label{eq:mm_agreement}
\end{equation}
By convexity of $x^2$, we have the global lower bound
\begin{equation}
x^2\ge 2x^{(t)}x-\left(x^{(t)}\right)^2 .
\label{eq:mm_lower_bound}
\end{equation}
Applying Eq.~\eqref{eq:mm_lower_bound} to every quadratic term yields the surrogate
\begin{equation}
\begin{aligned}
G^{(t)}(\mathbf{T},\Theta)=
&\sum_{m=1}^{M}2a_{v,m}^{(t)}
\left\langle \mathbf{T},\mathbf{P}^{(m)}(\Theta^{(m)})\right\rangle\\
&+\sum_{m=1}^{M}2a_{s,m}^{(t)}
\left\langle \mathbf{T},\hat{\mathbf{P}}^{(m)}\right\rangle
+\varepsilon H(\mathbf{T}),
\end{aligned}
\label{eq:mm_surrogate}
\end{equation}
up to additive constants independent of $\mathbf{T}$ and $\Theta$. We maximize this surrogate with respect to the transport plan under fixed visual posteriors, and then refresh the GMM parameters using the resulting soft assignments.

\paragraph{Update of the transport plan}
With $\Theta^{(m,t)}$ fixed, define the adaptive coefficients
\begin{equation}
\omega_{v,m}^{(t)}=2a_{v,m}^{(t)},\qquad
\omega_{s,m}^{(t)}=2a_{s,m}^{(t)},
\label{eq:mm_weights}
\end{equation}
and the fused score matrix
\begin{equation}
\begin{aligned}
\mathbf{S}^{(t)}=
\sum_{m=1}^{M}\omega_{v,m}^{(t)}
\mathbf{P}^{(m)}\!\left(\Theta^{(m,t)}\right)
+\sum_{m=1}^{M}\omega_{s,m}^{(t)}\hat{\mathbf{P}}^{(m)}.
\end{aligned}
\label{eq:score_matrix}
\end{equation}
The $\mathbf{T}$-subproblem becomes
\begin{equation}
\mathbf{T}^{(t+1)}=
\arg\max_{\mathbf{T}\in\Pi(\mathbf{r},\mathbf{q})}
\left\langle \mathbf{T},\mathbf{S}^{(t)}\right\rangle
+\varepsilon H(\mathbf{T}),
\label{eq:mm_update}
\end{equation}
which is a standard entropic OT problem and can be solved efficiently by Sinkhorn iterations \cite{OT}. Specifically, we form the elementwise kernel
\begin{equation}
\mathbf{K}^{(t)}=\exp\left(\mathbf{S}^{(t)}/\varepsilon\right),
\label{eq:kernel}
\end{equation}
and compute
\begin{equation}
\mathbf{T}^{(t+1)}=
\mathrm{Diag}\!\left(\mathbf{u}^{(t)}\right)
\mathbf{K}^{(t)}
\mathrm{Diag}\!\left(\mathbf{v}^{(t)}\right),
\label{eq:sinkhorn}
\end{equation}
where $\mathbf{u}^{(t)}$ and $\mathbf{v}^{(t)}$ are alternately rescaled so that the row and column marginals match $\mathbf{r}$ and $\mathbf{q}$.

\paragraph{Update of the visual GMMs}
With $\mathbf{T}^{(t+1)}$ fixed, we convert it into row-normalized soft assignments
\begin{equation}
\bar{\mathbf{T}}^{(t+1)}=\mathrm{Diag}(\mathbf{r})^{-1}\mathbf{T}^{(t+1)}.
\label{eq:tbar_iter}
\end{equation}
For each VLM $m$, we treat $\bar{\mathbf{T}}^{(t+1)}$ as soft assignments of samples to semantic categories and perform a moment update for the visual Gaussian mixture in its own embedding space. The component prior is fixed by the class marginal, i.e., $\pi_k^{(m)}=q_k$, which equals $1/C$ under the uniform $\mathbf{q}$ used in our experiments. The visual moment statistics are
\begin{equation}
\boldsymbol{\mu}_k^{(m,t+1)}=
\frac{\sum_{i=1}^{n}\bar{T}_{ik}^{(t+1)}\mathbf{z}_i^{(m)}}
{\sum_{i=1}^{n}\bar{T}_{ik}^{(t+1)}},
\label{eq:gmm_mu}
\end{equation}
\begin{equation}
\boldsymbol{\Sigma}^{(m,t+1)}
=\mathrm{Diag}\!\left(
\frac{1}{n}\sum_{i=1}^{n}\sum_{k=1}^{C}\bar{T}_{ik}^{(t+1)}
\left(\mathbf{z}_i^{(m)}-\boldsymbol{\mu}_k^{(m,t+1)}\right)^{\odot 2}
\right),
\label{eq:gmm_sigma}
\end{equation}
In implementation, we use a damped moment update during transport-conditioned GMM refinement. The undamped weighted moments in Eqs.~\eqref{eq:gmm_mu}--\eqref{eq:gmm_sigma} define the target GMM statistics induced by the current transport assignments, and the actual mean and covariance parameters are updated by a convex combination between the previous parameters and these target statistics. This relaxation stabilizes the alternating procedure under noisy early transport assignments. After this update, the visual posterior matrix $\mathbf{P}^{(m)}(\Theta^{(m,t+1)})$ is recomputed through Eq.~\eqref{eq:visual_prob}. With fixed visual posteriors, the transport update maximizes the MM surrogate induced by the tight lower bound in Eq.~\eqref{eq:mm_lower_bound}. The subsequent GMM update is a transport-conditioned moment refinement that treats $\bar{\mathbf{T}}^{(t+1)}$ as soft class assignments in each VLM feature space. In practice, we alternate these two updates for a small fixed number of iterations.

\subsection{One Stone, Three Birds}
After optimization, OSTB contains a consensus transport plan, refined visual GMMs, and branch-specific reliability coefficients. These quantities are not separate task-specific heuristics. They are different views of the same target-side latent structure and can be translated into three different deployment outputs: model selection, target-aware adaptation, and ensembling.

Let $\mathbf{T}^{\star}$ denote the converged transport plan and let
\begin{equation}
\bar{\mathbf{T}}=\mathrm{Diag}(\mathbf{r})^{-1}\mathbf{T}^{\star}
\label{eq:tbar}
\end{equation}
be its row-normalized form. Since $\bar{\mathbf{T}}\mathbf{1}_{C}=\mathbf{1}_{n}$, each row of $\bar{\mathbf{T}}$ gives a transport-induced soft assignment over the target label set. We do not report these in-sample assignments as final predictions; instead, they provide latent target supervision for estimating model reliability and building classifiers that can be applied to held-out target test samples.

\paragraph{Model selection}
The first output is a label-free reliability ranking over the candidate VLMs. Since $\mathbf{T}^{\star}$ summarizes the target structure supported by both semantic and visual branches, a candidate should be trusted when both of its branches agree with this consensus. At convergence, the MM procedure yields visual and semantic reliability coefficients $\{\omega_{v,m}^{\star}\}_{m=1}^{M}$ and $\{\omega_{s,m}^{\star}\}_{m=1}^{M}$ from Eq.~\eqref{eq:mm_weights}. We first normalize them within each branch:
\begin{equation}
\alpha_m=
\frac{\omega_{v,m}^{\star}}{\sum_{\ell=1}^{M}\omega_{v,\ell}^{\star}},
\qquad
\beta_m=
\frac{\omega_{s,m}^{\star}}{\sum_{\ell=1}^{M}\omega_{s,\ell}^{\star}}.
\label{eq:deployment_weights}
\end{equation}
Here $\alpha_m$ measures the reliability of the transport-conditioned visual GMM branch of model $m$, and $\beta_m$ measures the reliability of its semantic VLM branch. We use their sum as the model-selection score:
\begin{equation}
s_m^{\mathrm{sel}}=\alpha_m+\beta_m .
\label{eq:score}
\end{equation}
OSTB ranks candidate models in descending order of $s_m^{\mathrm{sel}}$, and the selected model is
\begin{equation}
m^{\star}=\arg\max_{m}s_m^{\mathrm{sel}}.
\label{eq:select}
\end{equation}
This score uses the same two reliability views that drive the final ensemble: $\beta_m$ preserves agreement with the label-aligned semantic branch, while $\alpha_m$ incorporates whether the model's visual representation supports a coherent transport-conditioned classifier. The resulting ranking therefore reflects compatibility with the shared target structure rather than semantic confidence alone.

\paragraph{Transport-conditioned adaptation}
The second output is target-aware adaptation. The row-normalized plan $\bar{\mathbf{T}}$ provides soft class assignments for the unlabeled target adaptation samples. Because its columns are aligned with the semantic prior $\mathcal{Y}$, these assignments give semantic identities to the visual clusters in each VLM feature space. After GMM refinement, the $k$-th Gaussian component of model $m$ no longer represents an anonymous feature cluster, but a target-aware visual component associated with text label $y_k$.

Given a held-out target test sample $x$, we extract the image embedding of the $m$-th VLM and evaluate the optimized Gaussian mixture:
\begin{equation}
\tilde{p}_{k}^{(m)}(x)=
\frac{\pi_k^{(m,\star)}
\mathcal{N}\!\left(\mathbf{z}^{(m)}(x)\mid
\boldsymbol{\mu}_k^{(m,\star)},\boldsymbol{\Sigma}^{(m,\star)}\right)}
{\sum_{j=1}^{C}\pi_j^{(m,\star)}
\mathcal{N}\!\left(\mathbf{z}^{(m)}(x)\mid
\boldsymbol{\mu}_j^{(m,\star)},\boldsymbol{\Sigma}^{(m,\star)}\right)},
\label{eq:adapt_prob}
\end{equation}
where $\Theta^{(m,\star)}=\{\{\pi_k^{(m,\star)},\boldsymbol{\mu}_k^{(m,\star)}\}_{k=1}^{C},\boldsymbol{\Sigma}^{(m,\star)}\}$ uses the fixed class-marginal priors and the visual statistics estimated from the unlabeled target adaptation set through Eqs.~\eqref{eq:gmm_mu}--\eqref{eq:gmm_sigma}. Stacking $\{\tilde{p}_{k}^{(m)}(x)\}_{k=1}^{C}$ gives the adapted posterior $\tilde{\mathbf{p}}^{(m)}(x)\in[0,1]^C$.

This adapted posterior acts as a visual correction branch for model $m$: it carries target-domain structure learned from the unlabeled target adaptation set while remaining indexed by the semantic label set. At deployment, OSTB also keeps the original zero-shot semantic branch. For the same held-out target test sample $x$, we compute $\hat{\mathbf{p}}^{(m)}(x)$ using the frozen image-text matching rule in Eq.~\eqref{eq:semantic_prob}. The two branches are complementary: the semantic branch preserves direct text-label alignment, while the adapted GMM branch contributes target-conditioned visual evidence.

\paragraph{Ensembling}
The third output is reliability-aware ensembling. The same normalized coefficients in Eq.~\eqref{eq:deployment_weights} are reused as deployment weights, with $\sum_{m=1}^{M}\alpha_m=1$ and $\sum_{m=1}^{M}\beta_m=1$. During optimization, these coefficients determine how strongly each branch shapes the shared transport consensus; after optimization, they determine how strongly each branch contributes to prediction. The final ensemble prediction is
\begin{equation}
\mathbf{p}_{\mathrm{ens}}(x)=
\sum_{m=1}^{M}\alpha_m\tilde{\mathbf{p}}^{(m)}(x)
+\sum_{m=1}^{M}\beta_m\hat{\mathbf{p}}^{(m)}(x).
\label{eq:adapted_ensemble}
\end{equation}
We normalize $\mathbf{p}_{\mathrm{ens}}(x)$ over classes when interpreting it as a probability distribution for prediction. For single-model deployment, OSTB selects the backbone through the joint reliability ranking in Eq.~\eqref{eq:score} and then combines the selected model's semantic branch with its transport-adapted GMM branch. For full multi-model deployment, OSTB uses Eq.~\eqref{eq:adapted_ensemble}. Thus, model ranking, target-domain adaptation, and ensembling are all derived from the same transport structure $\mathbf{T}^{\star}$ rather than designed as disconnected procedures.

\section{Experiments}\label{sec:experiments}
This section evaluates OSTB under the same coupled view used in the method: heterogeneous VLM deployment requires model selection, target adaptation, and ensembling to be assessed together. We therefore examine both the raw shifts in model preference across target domains and the gains delivered by transport-based integration.

\subsection{Experimental Settings}

\subsubsection{Evaluation Protocol}
We organize experiments over three application domains: natural images, remote sensing, and medical pathology. For each benchmark, the semantic prior in the form of target class names is available. During adaptation, OSTB only uses the unlabeled target adaptation split; labels from the disjoint target test split are used only for reporting evaluation metrics. The central empirical question is whether models with different pre-training histories exhibit dataset-dependent preferences, and whether OSTB can exploit this heterogeneity more reliably than heuristic backbone selection. We use adapted-classifier deployment as the main protocol. The unlabeled target adaptation split is first used to estimate the shared transport plan, transport-conditioned visual GMM classifiers, and model suitability weights. The resulting adapted classifiers and weights are then evaluated on the held-out target test split. The in-sample transport consensus only provides latent structure for adaptation and selection; it is not reported as the final classification output.

Our evaluation is centered on three complementary questions: \emph{First, can OSTB rank candidate VLMs in a way that is consistent with their actual target-domain utility? Second, does the hybrid adapted ensemble predictor outperform single-model deployment or one-branch evidence? Third, as the candidate pool becomes larger or more heterogeneous, can OSTB preserve reliable predictions instead of being dominated by less suitable models?}

\subsubsection{Datasets}
We substantially expand the evaluation suite across natural-image, remote-sensing, and medical-pathology scenarios. The natural-image benchmarks include ImageNet, SUN397, FGVC Aircraft, EuroSAT, Stanford Cars, Food101, Oxford Pets, Oxford Flowers, Caltech101, DTD, UCF101, CIFAR-10/100, and CUB-200-2011 \cite{imagenet,sun397,aircraft,eurosat,scars,food101,pets,flowers,caltech101,dtd,ucf101,cifar,cub}. The medical-pathology benchmarks include SICAP-MIL, PCam, Osteosarcoma, BACH, BreakHis, SkinCancer, SkinTumor, LC25000 Lung/Colon, NCT-CRC, WSSS4LUAD, and PanNuke \cite{sicap,pcam,osteosarcoma,bach,breakhis,skincancer,lung,nct,luad,gamper2020pannuke}. The remote-sensing benchmarks include AID, EuroSAT, MLRSNet, OPTIMAL31, PatternNet, RESISC45, RSC11, RSICB128/256, and WHURS19 \cite{AID,eurosat,MLRSNet,OPTIMAL,PatternNet,RESISC45,RSC11,RSICB,WHURS19}. This broader coverage tests whether the proposed selection-centered framework remains reliable as both target domains and candidate VLMs become more diverse. Tables~\ref{tab:natural_dataset_stats}--\ref{tab:remote_dataset_stats} summarize dataset statistics used in our experiments. Because several benchmarks retain official validation splits or unused remainder samples under our protocol, the adaptation and test counts do not necessarily sum to total sample count in every table entry.

\begin{table}[!t]
\caption{Natural-image benchmark statistics.}
\label{tab:natural_dataset_stats}
\centering
\scriptsize
\setlength{\tabcolsep}{4.2pt}
\renewcommand{\arraystretch}{1.05}
\resizebox{\columnwidth}{!}{%
\begin{tabular}{lrrrr}
\toprule
Dataset & Total Samples & Adapt & Test & Classes \\
\midrule
ImageNet & 1,331,167 & 1,281,167 & 50,000 & 1000 \\
SUN397 & 39,700 & 15,880 & 19,850 & 397 \\
FGVC Aircraft & 10,000 & 3,334 & 3,333 & 100 \\
EuroSAT & 27,000 & 13,500 & 8,100 & 10 \\
Stanford Cars & 16,185 & 6,509 & 8,041 & 196 \\
Food101 & 101,000 & 50,500 & 30,300 & 101 \\
Oxford Pets & 7,349 & 2,944 & 3,669 & 37 \\
Oxford Flowers & 8,189 & 4,093 & 2,463 & 102 \\
Caltech101 & 8,242 & 4,128 & 2,465 & 100 \\
DTD & 5,640 & 2,820 & 1,692 & 47 \\
UCF101 & 13,320 & 7,639 & 3,783 & 101 \\
CIFAR-10 & 60,000 & 40,000 & 10,000 & 10 \\
CIFAR-100 & 60,000 & 40,000 & 10,000 & 100 \\
CUB-200-2011 & 11,788 & 4,794 & 5,794 & 200 \\
\bottomrule
\end{tabular}%
}
\end{table}

\begin{table}[!t]
\caption{Medical-pathology benchmark statistics.}
\label{tab:pathology_dataset_stats}
\centering
\scriptsize
\setlength{\tabcolsep}{4.2pt}
\renewcommand{\arraystretch}{1.05}
\resizebox{\columnwidth}{!}{%
\begin{tabular}{lrrrr}
\toprule
Dataset & Total Samples & Adapt & Test & Classes \\
\midrule
SICAP-MIL & 1,501 & 1,050 & 451 & 4 \\
PCam & 327,680 & 262,144 & 32,768 & 2 \\
Osteosarcoma & 1,144 & 800 & 344 & 3 \\
BACH & 400 & 280 & 120 & 4 \\
BreakHis & 7,909 & 5,536 & 2,373 & 8 \\
SkinCancer & 129,364 & 88,971 & 28,039 & 16 \\
SkinTumor & 42,416 & 29,419 & 8,851 & 4 \\
LC25000 Lung & 15,000 & 10,500 & 4,500 & 3 \\
LC25000 Colon & 10,000 & 7,000 & 3,000 & 2 \\
NCT-CRC & 107,180 & 50,000 & 7,180 & 9 \\
WSSS4LUAD & 10,091 & 7,063 & 3,028 & 2 \\
PanNuke & 6,234 & 4,346 & 1,888 & 2 \\
\bottomrule
\end{tabular}%
}
\end{table}

\begin{table}[!t]
\caption{Remote-sensing benchmark statistics.}
\label{tab:remote_dataset_stats}
\centering
\scriptsize
\setlength{\tabcolsep}{4.2pt}
\renewcommand{\arraystretch}{1.05}
\resizebox{\columnwidth}{!}{%
\begin{tabular}{lrrrr}
\toprule
Dataset & Total Samples & Adapt & Test & Classes \\
\midrule
AID & 10,000 & 7,000 & 3,000 & 30 \\
EuroSAT & 27,000 & 18,900 & 8,100 & 10 \\
MLRSNet & 109,161 & 76,410 & 32,751 & 46 \\
OPTIMAL31 & 1,860 & 1,302 & 558 & 31 \\
PatternNet & 30,400 & 21,280 & 9,120 & 38 \\
RESISC45 & 31,500 & 22,050 & 9,450 & 45 \\
RSC11 & 1,232 & 862 & 370 & 11 \\
RSICB128 & 36,707 & 25,696 & 11,011 & 45 \\
RSICB256 & 24,747 & 17,323 & 7,424 & 35 \\
WHURS19 & 1,005 & 703 & 302 & 19 \\
\bottomrule
\end{tabular}%
}
\end{table}

\subsubsection{Models}
The candidate model pools are constructed in a domain-aware manner. The main experiments are performed separately within each application domain, so each pool reflects a realistic deployment scenario in which several plausible models with different pre-training histories are available for the same application.

\noindent\textbf{Natural images.}
For natural-image benchmarks, we use an expanded general-purpose VLM pool spanning several release generations of CLIP-style models. To avoid making model selection trivially correlated with model size, we retain B-scale or comparable variants whenever possible. The pool includes OpenAI CLIP ViT-B/32 and ViT-B/16 \cite{radford2021learning}, ALIGN \cite{jia2021scaling}, SLIP-ViT-B/16 \cite{mu2022slip}, OpenCLIP-ViT-B/16 \cite{cherti2023reproducible}, DeCLIP-ViT-B/32 \cite{li2022supervision}, FLAVA \cite{singh2022flava}, CoCa-ViT-B/32 \cite{yu2022coca}, EVA-CLIP-ViT-B/16 \cite{sun2023evaclip}, MetaCLIP-ViT-B/16 \cite{xu2024demystifying}, JinaCLIPv2 \cite{koukounas2024jinaclipv2}, SigLIP2-ViT-B/16 \cite{siglip}, PE-Core-ViT-B/16 \cite{bolya2025perception}, and MobileCLIP2-B \cite{faghri2025mobileclip2}.

\noindent\textbf{Remote sensing.}
For remote-sensing benchmarks, we combine general-purpose CLIP baselines with remote-sensing-oriented VLMs. The pool includes OpenAI CLIP ViT-B/16 and ViT-B/32 \cite{radford2021learning}, GeoRSCLIP-ViT-B/32 \cite{georsclip}, RemoteCLIP-ViT-B/32 \cite{remoteclip}, SkyCLIP50-ViT-B/32 \cite{skyscript}, RS-M-CLIP \cite{silva2024multilingual}, RSDiX-CLIP-ViT-B/16 and RSDiX-CLIP-ViT-B/32 \cite{terlizzi2025rsdix}, and StreetCLIP \cite{haas2023learning}. This setting reflects a realistic remote-sensing deployment scenario in which generic visual-language priors and geospatially specialized pre-training may both be available.

\noindent\textbf{Medical pathology.}
For computational pathology benchmarks, we use an expanded medical VLM pool ordered by release generation. The pool includes OpenAI CLIP ViT-B/32 and ViT-B/16 \cite{radford2021learning}, BiomedCLIP \cite{biomedclip}, PLIP \cite{pliphuang2023visual}, QuiltNet-ViT-B/16 and QuiltNet-ViT-B/32 \cite{ikezogwo2023quilt}, CONCH \cite{conchlu2024visual}, UniMed-CLIP \cite{khattak2024unimed}, KEEP \cite{zhou2024keep}, PMC-CLIP \cite{lin2023pmcclip}, PathGen-CLIP \cite{sun2025pathgen}, and MUSK \cite{muskxiang2025vision}.

\subsubsection{Implementation Details}
For every benchmark, we first construct a domain-consistent candidate pool and obtain VLM probability outputs and visual representations using the corresponding released checkpoints and official inference preprocessing pipelines. Each benchmark uses a single domain-appropriate prompt template, and we do not perform prompt ensembling or prompt optimization. This protocol keeps the comparison focused on target-domain deployment rather than additional prompt engineering.

For adapted-classifier deployment, an unlabeled target adaptation split is processed jointly. OSTB receives the original probability outputs and visual representations of all candidate VLMs on that split and produces a shared transport plan, branch-specific reliability weights, and transport-conditioned visual GMM classifiers. These learned GMMs and weights are then frozen and applied to held-out target test samples by fusing the adapted visual posteriors with the original semantic posteriors of the same VLMs through Eq.~\eqref{eq:adapted_ensemble}. This protocol tests whether deployment-time model choice and integration can be improved by exploiting the shared target structure revealed by multiple heterogeneous VLMs, without using target labels during adaptation.

Unless otherwise specified, OSTB uses $\epsilon=0.05$ and 15 alternating iterations. During transport-conditioned GMM refinement, we use a damped moment update for the GMM parameters. The update coefficient is larger in early iterations to quickly correct noisy initialization and gradually decreases as the transport assignments become more stable. This design keeps the moment-update form above while acting as a temporal regularizer that prevents transient noisy assignments from over-shifting the class components. All experiments are conducted on a single NVIDIA GeForce RTX 4090 GPU.

\subsection{Main Results}

We organize the main results according to the deployment logic of OSTB. We first evaluate whether the method provides a reliable ranking over candidate VLMs. We then isolate the adaptation effect at the single-model level to test whether the transport-conditioned GMM branch improves target-domain prediction. Finally, we evaluate full multi-model integration to determine whether combining adapted candidates yields gains beyond ranking and single-model adaptation.

\noindent\textbf{1) Model selection results.} We first evaluate whether OSTB can provide a reliable ranking over candidate VLMs. For analysis, we compare it with several label-free baselines, including KL divergence, cross-entropy, entropy, and confidence. The oracle ranking for each dataset is defined by the zero-shot Top-1 accuracy of candidate VLMs on the held-out target test split. Table~\ref{tab:selection_ablation} reports the overall average over 36 benchmarks. Spearman's $\rho$ and Kendall's $\tau$ measure full-ranking agreement, \textit{Hit@1} measures whether the top-ranked model coincides with the oracle best model, \textit{Top-3} checks whether the oracle best model appears within the predicted top three, \textit{Regret} is the zero-shot accuracy gap between the oracle best model and the top-ranked model, and \textit{Sel.Acc} is the average zero-shot accuracy of the top-ranked model. We also report BestModel as an oracle selected-accuracy upper bound.

\begin{table}[!t]
\caption{Overall model-selection results averaged over 36 benchmarks. Ranking correlations, hit rates, regret, and selected-model accuracy are computed against the oracle ranking induced by held-out target test accuracy. Unless otherwise specified, bold and underline denote the best and second-best results, respectively. Diagnostic GMM-only rows in the main accuracy tables are excluded from this highlighting convention.}
\label{tab:selection_ablation}
\centering
\scriptsize
\setlength{\tabcolsep}{5.0pt}
\renewcommand{\arraystretch}{1.08}
\resizebox{\columnwidth}{!}{%
\begin{tabular}{lcccccc}
\toprule
Criterion & $\rho\uparrow$ & $\tau\uparrow$ & Hit@1$\uparrow$ & Top-3$\uparrow$ & Regret$\downarrow$ & Sel.Acc$\uparrow$ \\
\midrule
BestModel & -- & -- & -- & -- & -- & \textbf{78.28} \\
OSTB & \textbf{0.807} & \textbf{0.680} & \textbf{0.667} & \textbf{0.917} & \textbf{1.249} & \underline{77.03} \\
KL Divergence & 0.705 & 0.619 & \underline{0.528} & \underline{0.667} & 3.058 & 75.22 \\
Cross-Entropy & \underline{0.711} & \underline{0.623} & 0.500 & \underline{0.667} & \underline{3.029} & 75.25 \\
Entropy & 0.386 & 0.305 & 0.194 & 0.639 & 10.242 & 68.04 \\
Confidence & 0.436 & 0.355 & 0.278 & \underline{0.667} & 9.875 & 68.40 \\
\bottomrule
\end{tabular}
}
\end{table}

The overall selection results show that OSTB provides the most reliable label-free ranking among the compared criteria. It achieves the best full-ranking correlations, with Spearman's $\rho$ of 0.807 and Kendall's $\tau$ of 0.680, and it also gives the strongest label-free deployment choice, with the best Hit@1 (0.667), Top-3 rate (0.917), lowest regret (1.249), and selected-model accuracy of 77.03\%. The oracle BestModel upper bound reaches 78.28\%, so OSTB remains within 1.25 percentage points of selecting the dataset-wise best candidate with test-label access. The Hit@1 value is not saturated, but this does not imply unstable deployment: on several benchmarks, multiple candidate VLMs have similar zero-shot performance, so small score differences can change the oracle top-1 identity even when the predicted top model remains highly competitive. The Top-3 rate of 0.917 is therefore particularly informative, showing that OSTB usually places a strong candidate within a small shortlist. Among the non-OSTB criteria, cross-entropy is the strongest competing baseline on full-ranking correlation and selected-model accuracy, while KL divergence is competitive for top-1 deployment. Entropy and confidence are substantially weaker across the reported metrics.

\begin{figure*}[!t]
\centering
\includegraphics[width=1\textwidth]{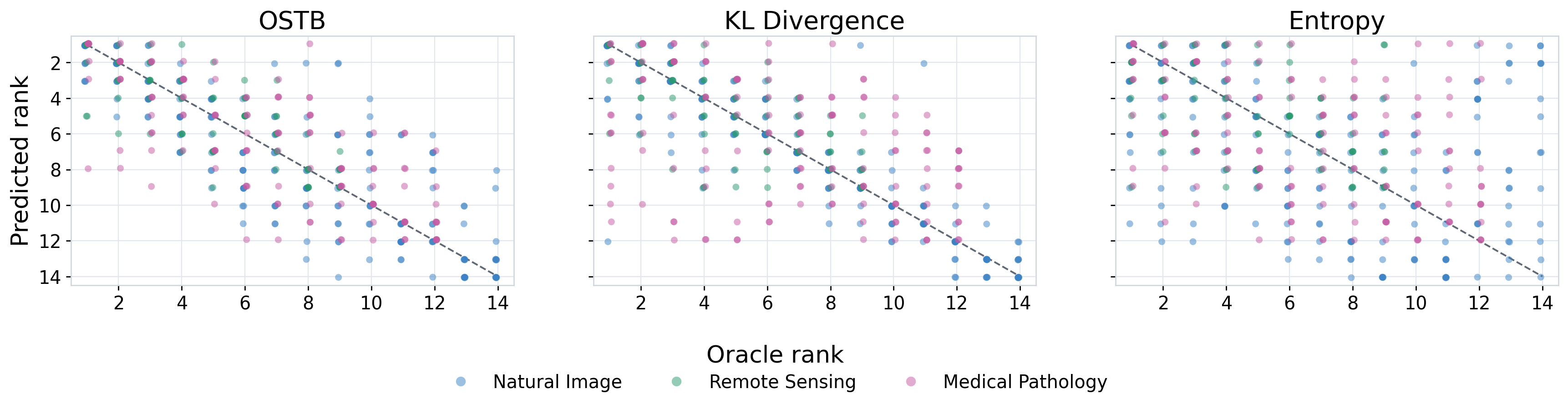}
\caption{Predicted model ranking versus oracle ranking. Each point denotes a candidate VLM on one benchmark, with the horizontal axis showing the oracle rank induced by zero-shot target accuracy and the vertical axis showing the predicted rank. Points closer to the diagonal indicate more faithful recovery of the global ordering.}
\label{fig:ranking_vs_oracle}
\end{figure*}

Figure~\ref{fig:ranking_vs_oracle} provides a visual comparison of the predicted and oracle rankings. The OSTB points form a more compact cloud around the diagonal trend than the competing label-free criteria, suggesting better overall ranking consistency rather than only isolated top-rank matches. In contrast, KL divergence and entropy show more dispersed patterns, indicating weaker agreement with the oracle ordering. Together, the table and figure indicate that OSTB provides a more stable reliability ranking before any later deployment choice is made.

\begin{table}[!t]
\caption{Ranking-guided Top-$k$ deployment accuracy (\%). The selected zero-shot row reports the top-ranked original VLM, while the Top-$k$ and All rows use OSTB-ranked candidate subsets under the adapted-classifier deployment protocol.}
\label{tab:topk_deployment}
\centering
\scriptsize
\setlength{\tabcolsep}{5.0pt}
\renewcommand{\arraystretch}{1.08}
\begin{tabular}{lcccc}
\toprule
Candidate Pool & Natural & Remote & Pathology & Overall \\
\midrule
Selected zero-shot & 81.49 & 75.12 & 73.41 & 77.03 \\
Top-1 & 84.62 & 79.95 & 76.62 & 80.66 \\
Top-2 & \underline{85.59} & 83.38 & \textbf{78.39} & 82.57 \\
Top-4 & \textbf{85.83} & \textbf{86.24} & 78.25 & \textbf{83.42} \\
All & 84.68 & \underline{85.37} & \underline{78.33} & \underline{82.75} \\
\bottomrule
\end{tabular}
\end{table}

The ranking can also prune the candidate pool before full deployment. As shown in Table~\ref{tab:topk_deployment}, retaining only the top-ranked models already improves the overall accuracy from 77.03\% for the selected zero-shot baseline to 80.66\% with Top-1, and further to 82.57\% and 83.42\% with Top-2 and Top-4. The gain is consistent across all three application domains, indicating that the OSTB ranking is useful not only for analyzing oracle preference but also for constructing compact deployment subsets. Compared with using all candidates, Top-4 achieves higher natural-image, remote-sensing, and overall averages, while the full pool remains slightly stronger on medical pathology. This pattern suggests that most useful model diversity is concentrated in a small set of highly ranked candidates; adding more models can still help on some individual datasets, but it may also introduce less reliable evidence that offsets part of the ensemble benefit. Ranking-guided pruning is therefore useful when latency, memory, or model availability makes evaluating every candidate impractical.

\noindent\textbf{2) Model adaptation results.} We next isolate the adaptation effect before final semantic-visual fusion. Table~\ref{tab:single_model_adaptation} compares three predictors averaged over candidate VLMs: the original VLM probability output, a GMM branch adapted independently for each VLM, and a GMM branch adapted inside the full multi-VLM OSTB optimization. This comparison tests whether sharing target structure across multiple VLMs provides stronger adaptation supervision than adapting each model in isolation.

\begin{table}[!t]
\caption{Model adaptation results averaged over candidate VLMs and benchmarks. The table compares original zero-shot probability outputs, independently adapted single-VLM GMM branches, and multi-VLM GMM branches refined through the shared OSTB transport structure.}
\label{tab:single_model_adaptation}
\centering
\scriptsize
\setlength{\tabcolsep}{5.0pt}
\renewcommand{\arraystretch}{1.08}
\begin{tabular}{lcccc}
\toprule
Method & Natural & Remote & Pathology & Overall \\
\midrule
Zero-shot & 65.19 & 65.03 & 54.07 & 61.43 \\
Single-VLM GMM & \underline{67.13} & \underline{74.44} & \underline{62.63} & \underline{67.15} \\
Multi-VLM GMM & \textbf{73.86} & \textbf{81.60} & \textbf{73.20} & \textbf{75.26} \\
\bottomrule
\end{tabular}
\end{table}

Table~\ref{tab:single_model_adaptation} shows that isolated adaptation already improves the zero-shot VLM probability output, raising the overall average from 61.43\% to 67.15\%. More importantly, multi-VLM collaborative adaptation is substantially stronger than single-model adaptation. The multi-VLM GMM reaches 75.26\% overall, exceeding the single-VLM GMM by 8.11 points. The gain is consistent across domains: 73.86\% versus 67.13\% on natural images, 81.60\% versus 74.44\% on remote sensing, and 73.20\% versus 62.63\% on medical pathology. These results indicate that the shared transport structure does not merely support the final ensemble; it also provides better latent supervision for refining each model's visual GMM branch. Dataset-specific results for the adapted branches are further reported in Tables~\ref{tab:natural_results}--\ref{tab:pathology_results}.

\noindent\textbf{3) Multi-model integration results.} We finally evaluate the complete OSTB predictor. Tables~\ref{tab:natural_results}--\ref{tab:pathology_results} compare the final OSTB ensemble with all original candidate VLMs under the same adapted-classifier deployment protocol. The tables also include single-VLM and multi-VLM GMM rows as diagnostic branch outputs to expose each candidate's adaptation behavior, but the following deployment comparison focuses on the original VLM rows and the final OSTB ensemble.

\begin{table*}[!t]
	\caption{Top-1 Accuracy (\%) on natural-image benchmarks. Rows report original zero-shot VLM outputs, diagnostic single-VLM and multi-VLM GMM branches, and the final OSTB adapted ensemble evaluated on held-out target test samples.}
	\label{tab:natural_results}
	\centering
	\scriptsize
	\renewcommand{\arraystretch}{1.08}
	\setlength{\tabcolsep}{5pt}
	\resizebox{\textwidth}{!}{%
		\begin{tabular}{lccccccccccccccc}
			\toprule
			Model & ImageNet & SUN397 & Aircraft & EuroSAT & StanfordCars & Food101 & Pets & Flower102 & Caltech101 & DTD & UCF101 & CIFAR-10 & CIFAR-100 & CUB & Avg. \\
			\midrule
			CLIP-B/32 & 62.01 & 62.06 & 19.17 & 45.27 & 60.09 & 80.39 & 87.44 & 66.67 & 91.40 & 42.85 & 63.44 & 88.32 & 64.37 & 52.66 & 63.30 \\
			\quad +multi-VLM GMM & 62.86 & 69.57 & 25.23 & 77.11 & 72.13 & 78.59 & 85.12 & 81.65 & 86.33 & 58.10 & 72.48 & 89.93 & 67.55 & 67.62 & 71.02 \\
			\quad +single-VLM GMM & 59.68 & 65.74 & 19.23 & 53.86 & 59.93 & 78.45 & 85.20 & 72.72 & 83.12 & 51.06 & 66.32 & 89.61 & 63.93 & 53.78 & 64.47 \\
			\midrule
			CLIP-B/16 & 66.74 & 62.59 & 24.69 & 48.36 & 65.55 & 85.87 & 89.10 & 70.69 & 93.31 & 43.32 & 67.49 & 90.08 & 68.30 & 54.99 & 66.51 \\
			\quad +multi-VLM GMM & 68.37 & 71.70 & 30.90 & 77.16 & 79.17 & 84.73 & 89.51 & 84.25 & 88.64 & 60.70 & 75.28 & 92.97 & 70.92 & 75.23 & 74.97 \\
			\quad +single-VLM GMM & 66.05 & 67.79 & 24.30 & 63.25 & 67.48 & 84.89 & 88.72 & 77.87 & 87.55 & 48.64 & 71.61 & 92.99 & 67.91 & 60.72 & 69.27 \\
			\midrule
			ALIGN & 65.26 & 70.11 & 11.31 & 32.69 & 72.38 & 79.91 & 84.30 & 60.29 & 94.77 & 58.16 & 64.79 & 75.61 & 52.16 & 38.25 & 61.43 \\
			\quad +multi-VLM GMM & 67.04 & 72.42 & 24.27 & 73.49 & 83.16 & 80.74 & 80.38 & 81.49 & 88.72 & 63.83 & 72.30 & 81.14 & 57.71 & 61.25 & 70.57 \\
			\quad +single-VLM GMM & 63.71 & 71.68 & 13.08 & 33.58 & 74.97 & 79.01 & 78.14 & 66.02 & 89.98 & 58.87 & 70.00 & 79.00 & 49.80 & 38.94 & 61.91 \\
			\midrule
			SLIP-B/16 & 42.22 & 49.05 & 7.98 & 17.14 & 8.63 & 61.07 & 34.86 & 65.65 & 81.91 & 25.59 & 38.75 & 77.42 & 46.75 & 36.78 & 42.41 \\
			\quad +multi-VLM GMM & 57.54 & 67.66 & 20.49 & 82.58 & 28.09 & 70.57 & 63.83 & 83.68 & 83.29 & 58.92 & 69.57 & 85.82 & 62.26 & 60.18 & 63.89 \\
			\quad +single-VLM GMM & 44.57 & 56.03 & 9.36 & 33.04 & 6.68 & 65.81 & 39.71 & 71.58 & 73.51 & 35.64 & 46.74 & 83.13 & 52.00 & 37.45 & 46.80 \\
			\midrule
			OpenCLIP-B/16 & 72.98 & 69.88 & 29.73 & 56.37 & 89.90 & 87.52 & 92.83 & 75.44 & 96.67 & 58.33 & 67.46 & 96.27 & 81.99 & 77.53 & 75.21 \\
			\quad +multi-VLM GMM & 71.68 & 72.89 & 35.61 & 83.10 & 91.03 & 86.39 & 91.01 & 85.95 & 88.68 & 66.43 & 74.36 & 97.43 & 83.21 & 79.46 & 79.09 \\
			\quad +single-VLM GMM & 71.67 & 71.96 & 31.59 & 62.26 & 90.66 & 86.54 & 90.98 & 84.08 & 89.21 & 63.30 & 72.56 & 97.48 & 83.00 & 78.82 & 76.72 \\
			\midrule
			DeCLIP-B/32 & 39.45 & 47.77 & 2.70 & 24.20 & 3.58 & 46.79 & 34.89 & 62.32 & 77.57 & 25.41 & 36.32 & 80.88 & 49.94 & 34.98 & 40.49 \\
			\quad +multi-VLM GMM & 52.26 & 64.45 & 13.41 & 78.47 & 19.28 & 61.70 & 58.71 & 83.52 & 82.64 & 54.79 & 62.68 & 85.58 & 61.57 & 62.56 & 60.11 \\
			\quad +single-VLM GMM & 40.68 & 51.91 & 3.81 & 31.67 & 3.10 & 51.08 & 34.01 & 69.87 & 70.39 & 32.74 & 39.78 & 83.14 & 53.41 & 39.26 & 43.20 \\
			\midrule
			FLAVA & 54.89 & 59.10 & 12.09 & 37.43 & 27.48 & 74.74 & 68.85 & 58.83 & 90.67 & 38.71 & 52.42 & 89.90 & 64.45 & 46.53 & 55.44 \\
			\quad +multi-VLM GMM & 61.41 & 71.44 & 21.36 & 82.58 & 54.62 & 79.42 & 70.21 & 82.14 & 87.71 & 57.80 & 68.49 & 92.44 & 70.08 & 64.26 & 68.85 \\
			\quad +single-VLM GMM & 54.82 & 67.54 & 11.13 & 55.69 & 28.17 & 78.08 & 66.53 & 71.78 & 81.38 & 41.25 & 57.26 & 91.39 & 65.87 & 47.19 & 58.43 \\
			\midrule
			CoCa-B/32 & 63.43 & 66.59 & 18.06 & 45.43 & 86.42 & 75.44 & 89.13 & 65.21 & 94.52 & 52.48 & 62.25 & 93.55 & 73.85 & 60.27 & 67.62 \\
			\quad +multi-VLM GMM & 64.19 & 70.11 & 29.55 & 79.11 & 87.32 & 75.28 & 87.24 & 81.20 & 89.05 & 63.89 & 72.01 & 95.11 & 76.66 & 67.69 & 74.17 \\
			\quad +single-VLM GMM & 62.30 & 68.45 & 19.50 & 65.26 & 85.79 & 74.55 & 87.57 & 69.91 & 89.29 & 56.56 & 66.51 & 94.96 & 74.88 & 59.34 & 69.63 \\
			\midrule
			EVA-CLIP-B/16 & 74.47 & 70.81 & 24.69 & 58.17 & 79.13 & 86.55 & 92.20 & 75.80 & 97.16 & 50.35 & 69.13 & \textbf{98.26} & \underline{87.84} & 61.94 & 73.32 \\
			\quad +multi-VLM GMM & 73.15 & 72.66 & 35.34 & 81.53 & 85.97 & 85.83 & 91.44 & 86.36 & 91.52 & 63.00 & 75.28 & 98.74 & 87.75 & 80.39 & 79.21 \\
			\quad +single-VLM GMM & 72.20 & 71.29 & 27.39 & 75.52 & 81.63 & 85.65 & 91.33 & 79.29 & 91.28 & 53.96 & 72.67 & 98.73 & 88.03 & 70.16 & 75.65 \\
			\midrule
			MetaCLIP-B/16 & 69.97 & 68.81 & 28.62 & 55.20 & 74.69 & 84.07 & 90.49 & 73.81 & 96.11 & 51.77 & 68.04 & 89.87 & 64.95 & 68.29 & 70.34 \\
			\quad +multi-VLM GMM & 69.58 & 71.98 & 37.86 & 81.79 & 83.75 & 83.44 & 86.35 & 85.14 & 88.52 & 63.89 & 72.43 & 90.89 & 70.33 & 79.70 & 76.12 \\
			\quad +single-VLM GMM & 68.52 & 70.41 & 31.77 & 60.10 & 78.40 & 83.45 & 86.43 & 79.90 & 86.86 & 57.21 & 73.41 & 90.65 & 67.07 & 73.87 & 72.00 \\
			\midrule
			JinaCLIPv2 & 66.23 & 65.58 & 16.47 & 51.53 & 77.69 & 84.18 & 80.08 & 68.53 & 92.01 & 50.30 & 56.44 & 95.73 & 77.65 & 32.43 & 65.35 \\
			\quad +multi-VLM GMM & 75.73 & 75.78 & 31.80 & 76.89 & 84.45 & 88.68 & 87.49 & 85.83 & 89.01 & 63.89 & 74.70 & 96.56 & 79.86 & 66.29 & 76.93 \\
			\quad +single-VLM GMM & 70.16 & 71.10 & 18.42 & 56.19 & 76.76 & 88.11 & 83.73 & 71.42 & 85.92 & 52.54 & 58.63 & 96.55 & 78.58 & 36.99 & 67.51 \\
			\midrule
			SigLIP2-B/16 & 76.79 & 72.64 & 40.35 & 45.35 & 92.99 & \underline{90.01} & \underline{94.71} & 84.53 & \underline{97.53} & 62.94 & 73.09 & 94.05 & 72.95 & 50.19 & 74.87 \\
			\quad +multi-VLM GMM & 75.86 & 75.67 & 49.02 & 74.49 & 93.41 & 88.59 & 92.75 & 86.40 & 89.57 & 66.96 & 76.74 & 94.28 & 74.87 & 75.27 & 79.56 \\
			\quad +single-VLM GMM & 75.62 & 75.12 & 38.46 & 59.88 & 93.00 & 88.61 & 92.64 & 87.21 & 89.61 & 61.17 & 78.01 & 93.81 & 72.31 & 55.57 & 75.79 \\
			\midrule
			PE-Core-B/16 & \underline{78.38} & \underline{73.93} & \textbf{57.07} & \underline{61.74} & 93.43 & 89.83 & 94.66 & \textbf{87.01} & 97.48 & \underline{64.42} & \underline{79.17} & 97.14 & 82.96 & \underline{84.10} & \underline{81.52} \\
			\quad +multi-VLM GMM & 74.51 & 75.00 & 45.36 & 82.25 & 91.95 & 87.27 & 93.27 & 86.24 & 90.59 & 65.48 & 78.32 & 97.72 & 83.63 & 82.38 & 81.00 \\
			\quad +single-VLM GMM & 74.70 & 74.86 & 53.50 & 76.70 & 91.66 & 87.19 & 93.40 & 91.35 & 93.06 & 66.08 & 83.06 & 97.61 & 83.62 & 82.79 & 82.11 \\
			\midrule
			MobileCLIP2-B & 77.80 & 73.26 & 28.92 & 36.02 & \underline{93.66} & 88.31 & 93.38 & 84.61 & \textbf{97.85} & 61.11 & 68.68 & 92.79 & 72.11 & 80.13 & 74.90 \\
			\quad +multi-VLM GMM & 75.67 & 75.35 & 37.41 & 73.75 & 93.91 & 86.97 & 93.35 & 86.60 & 89.90 & 65.13 & 75.42 & 93.17 & 73.92 & 80.01 & 78.61 \\
			\quad +single-VLM GMM & 76.13 & 74.81 & 30.09 & 53.05 & 93.81 & 86.84 & 93.51 & 88.31 & 92.05 & 64.83 & 71.93 & 92.82 & 71.15 & 78.68 & 76.29 \\
			\midrule
			OSTB & \textbf{80.47} & \textbf{79.12} & \underline{52.24} & \textbf{86.49} & \textbf{94.96} & \textbf{90.46} & \textbf{95.20} & \underline{86.97} & 97.36 & \textbf{70.21} & \textbf{80.28} & \underline{98.25} & \textbf{88.38} & \textbf{85.05} & \textbf{84.68} \\
			\bottomrule
		\end{tabular}%
	}
\end{table*}

\noindent\textbf{Natural-image integration.} Table~\ref{tab:natural_results} shows that OSTB reaches an average accuracy of 84.68\%, outperforming the strongest original candidate VLM, PE-Core-B/16, at 81.52\%. In the main deployment comparison, it gives the best result on 10 of the 14 datasets, with especially clear gains on EuroSAT, SUN397, and DTD. The GMM-only diagnostic rows support the observation that multi-VLM adaptation usually improves the visual branch, while the final OSTB predictor combines semantic and visual evidence more reliably than any single original candidate VLM. OSTB is not the best deployment row on several datasets, including Aircraft, Flower102, Caltech101, and CIFAR-10; Aircraft shows the clearest preference for one particular model, while the remaining gaps are relatively small.

\begin{table*}[!t]
\caption{Top-1 Accuracy (\%) on remote-sensing benchmarks. Rows report original zero-shot VLM outputs, diagnostic single-VLM and multi-VLM GMM branches, and the final OSTB adapted ensemble evaluated on held-out target test samples.}
\label{tab:remote_results}
\centering
\scriptsize
\setlength{\tabcolsep}{3.2pt}
\renewcommand{\arraystretch}{1.08}
\resizebox{\textwidth}{!}{%
\begin{tabular}{lccccccccccc}
\toprule
Model & AID & EuroSAT & MLRSNet & OPTIMAL31 & PatternNet & RESISC45 & RSC11 & RSICB128 & RSICB256 & WHURS19 & Avg. \\
\midrule
CLIP-B/16 & 67.47 & 53.04 & 56.74 & 75.45 & 65.35 & 64.01 & 61.35 & 28.40 & 38.04 & 81.79 & 59.16 \\
\quad +multi-VLM GMM & 94.03 & 64.80 & 77.86 & 92.47 & 95.22 & 86.88 & 89.19 & 43.67 & 58.03 & 96.03 & 79.82 \\
\quad +single-VLM GMM & 78.57 & 67.52 & 64.49 & 87.63 & 87.58 & 74.51 & 82.16 & 33.22 & 42.17 & 95.70 & 71.35 \\
\midrule
CLIP-B/32 & 65.63 & 49.65 & 51.05 & 74.73 & 59.23 & 60.69 & 57.03 & 27.12 & 41.11 & 82.12 & 56.84 \\
\quad +multi-VLM GMM & 92.43 & 64.86 & 75.83 & 90.86 & 93.89 & 83.28 & 85.95 & 41.40 & 57.64 & 96.36 & 78.25 \\
\quad +single-VLM GMM & 81.97 & 45.77 & 63.67 & 85.48 & 78.37 & 73.10 & 60.00 & 34.26 & 43.93 & 95.36 & 66.19 \\
\midrule
GeoRSCLIP-B/32 & 70.77 & 47.06 & 64.90 & 79.03 & \underline{75.98} & 68.49 & 68.65 & 28.84 & 46.27 & 87.09 & 63.71 \\
\quad +multi-VLM GMM & 95.10 & 70.28 & 80.91 & 94.44 & 96.40 & 88.32 & 88.92 & 45.18 & 63.40 & 98.01 & 82.10 \\
\quad +single-VLM GMM & 78.50 & 56.56 & 70.93 & 87.10 & 95.42 & 76.48 & 83.24 & 44.76 & 62.31 & 98.01 & 75.33 \\
\midrule
RemoteCLIP-B/32 & 91.43 & 36.69 & 56.97 & 77.60 & 59.30 & 68.04 & 60.81 & 25.90 & 41.24 & 93.38 & 61.14 \\
\quad +multi-VLM GMM & 95.47 & 63.47 & 77.70 & 91.76 & 96.00 & 86.68 & 86.76 & 45.10 & 62.74 & 96.69 & 80.24 \\
\quad +single-VLM GMM & 95.03 & 47.35 & 67.84 & 90.14 & 78.36 & 81.28 & 72.43 & 32.08 & 48.18 & 97.02 & 70.97 \\
\midrule
SkyCLIP50-B/32 & 70.40 & 55.22 & 63.30 & 79.57 & 74.20 & 66.57 & 62.43 & \underline{39.40} & 47.64 & 91.39 & 65.01 \\
\quad +multi-VLM GMM & 93.90 & 68.70 & 80.61 & 92.65 & 96.17 & 86.33 & 88.38 & 46.64 & 61.31 & 97.02 & 81.17 \\
\quad +single-VLM GMM & 80.57 & 64.15 & 73.48 & 86.92 & 94.34 & 77.10 & 78.38 & 45.24 & 55.04 & 97.02 & 75.22 \\
\midrule
RS-M-CLIP & 92.60 & 33.15 & 63.96 & 98.39 & 52.20 & 95.82 & 64.59 & 33.83 & 44.64 & 95.70 & 67.49 \\
\quad +multi-VLM GMM & 98.43 & 67.48 & 82.91 & 100.00 & 96.21 & 99.08 & 90.27 & 47.08 & 63.74 & 98.01 & 84.32 \\
\quad +single-VLM GMM & 98.33 & 50.74 & 74.04 & 100.00 & 65.70 & 99.06 & 86.22 & 46.13 & 55.39 & 98.01 & 77.36 \\
\midrule
RSDiX-CLIP-B/16 & \underline{94.23} & 55.16 & \underline{73.48} & \underline{98.75} & 67.34 & \underline{97.83} & \underline{80.81} & 35.84 & \underline{49.42} & \underline{98.34} & \underline{75.12} \\
\quad +multi-VLM GMM & 97.97 & 63.72 & 82.48 & 99.10 & 96.32 & 98.10 & 92.70 & 46.86 & 63.44 & 99.67 & 84.04 \\
\quad +single-VLM GMM & 97.80 & 56.78 & 80.12 & 99.10 & 82.89 & 98.11 & 93.24 & 40.04 & 56.83 & 99.67 & 80.46 \\
\midrule
RSDiX-CLIP-B/32 & 92.03 & 49.56 & 67.02 & 96.95 & 60.67 & 97.04 & 80.27 & 32.97 & 46.94 & 96.03 & 71.95 \\
\quad +multi-VLM GMM & 97.13 & 69.01 & 81.17 & 97.67 & 95.56 & 98.33 & 92.16 & 45.95 & 62.06 & 99.01 & 83.81 \\
\quad +single-VLM GMM & 96.93 & 71.73 & 75.98 & 97.67 & 78.83 & 98.34 & 93.24 & 32.38 & 52.56 & 99.01 & 79.67 \\
\midrule
StreetCLIP & 71.77 & \underline{56.91} & 63.54 & 81.90 & 74.45 & 69.61 & 65.68 & 35.46 & 45.27 & 84.11 & 64.87 \\
\quad +multi-VLM GMM & 92.77 & 69.17 & 79.64 & 93.01 & 95.60 & 87.84 & 89.19 & 43.29 & 58.86 & 97.02 & 80.64 \\
\quad +single-VLM GMM & 80.17 & 66.44 & 69.08 & 90.14 & 87.92 & 79.45 & 78.65 & 39.14 & 46.50 & 96.69 & 73.42 \\
\midrule
OSTB & \textbf{98.70} & \textbf{70.70} & \textbf{85.22} & \textbf{99.28} & \textbf{96.72} & \textbf{98.80} & \textbf{89.46} & \textbf{50.53} & \textbf{64.59} & \textbf{99.67} & \textbf{85.37} \\
\bottomrule\end{tabular}%
}
\end{table*}

\noindent\textbf{Remote-sensing integration.} Table~\ref{tab:remote_results} gives the most consistent dataset-wise result. OSTB reaches an average accuracy of 85.37\%, compared with 75.12\% for the strongest original candidate, RSDiX-CLIP-B/16. In the main deployment comparison, it obtains the best result on all 10 remote-sensing benchmarks, showing that reliability-aware integration remains effective even when several domain-specific VLMs are available. The gains are especially large on EuroSAT, MLRSNet, and RSICB256, while the margins are smaller on benchmarks where individual remote-sensing VLMs are already strong. The adapted GMM rows further show that shared multi-VLM adaptation strengthens the visual branch, but the key observation is that the final semantic-visual ensemble delivers the strongest and most stable performance among the deployment rows.

\begin{table*}[!t]
\caption{Top-1 Accuracy (\%) on medical-pathology benchmarks. Rows report original zero-shot VLM outputs, diagnostic single-VLM and multi-VLM GMM branches, and the final OSTB adapted ensemble evaluated on held-out target test samples.}
\label{tab:pathology_results}
\centering
\scriptsize
\setlength{\tabcolsep}{2.8pt}
\renewcommand{\arraystretch}{1.08}
\begin{tabular}{lccccccccccccc}
\toprule
\multicolumn{1}{c}{Model} & 
 SICAP-MIL & 
 SKINCANCER & 
 SKINTUMOR & 
 LC-LUNG & 
 LC-COLON & 
 Osteo & 
 NCT-CRC & 
 WSSS4LUAD & 
 PanNuke & 
 PCam & 
 BACH & 
 BREAKHIS &
\multicolumn{1}{c}{Avg.} 
\\
\midrule
CLIP-B/32 & 34.59 & 2.70 & 17.23 & 43.60 & 51.07 & 59.01 & 24.14 & 64.89 & 62.13 & 54.77 & 25.00 & 18.16 & 38.11 \\
\quad +multi-VLM GMM & 46.56 & 59.65 & 69.71 & 88.82 & 93.63 & 76.16 & 75.31 & 77.51 & 72.30 & 50.67 & 50.00 & 25.71 & 65.50 \\
\quad +single-VLM GMM & 16.85 & 4.57 & 22.64 & 79.27 & 86.23 & 45.64 & 50.49 & 62.85 & 62.87 & 57.74 & 16.67 & 12.18 & 43.17 \\
\midrule
CLIP-B/16 & 29.71 & 3.54 & 14.52 & 30.98 & 50.43 & 48.84 & 24.28 & 64.89 & 49.58 & 47.13 & 40.83 & 7.71 & 34.37 \\
\quad +multi-VLM GMM & 49.67 & 58.32 & 66.55 & 89.87 & 92.20 & 79.07 & 76.39 & 79.52 & 75.79 & 49.61 & 52.50 & 24.36 & 66.15 \\
\quad +single-VLM GMM & 15.52 & 14.56 & 21.88 & 45.53 & 84.30 & 61.05 & 29.14 & 70.54 & 72.25 & 50.79 & 37.50 & 14.62 & 43.14 \\
\midrule
BiomedCLIP & 44.12 & 16.98 & 46.04 & 67.67 & 78.60 & 71.22 & 42.92 & 77.01 & 58.69 & 61.14 & 49.17 & 18.58 & 52.68 \\
\quad +multi-VLM GMM & 51.88 & 65.20 & 73.54 & 92.27 & 97.83 & 78.78 & 90.64 & 89.10 & 77.91 & 52.45 & 62.50 & 24.74 & 71.40 \\
\quad +single-VLM GMM & 45.90 & 34.48 & 63.45 & 86.33 & 96.70 & 66.57 & 74.26 & 87.85 & 73.62 & 63.70 & 48.33 & 19.89 & 63.42 \\
\midrule
PLIP & \underline{47.67} & 22.35 & 61.36 & 85.44 & 70.83 & 56.40 & 62.99 & 73.12 & 58.47 & 55.19 & 35.00 & 13.91 & 53.56 \\
\quad +multi-VLM GMM & 54.77 & 66.63 & 75.20 & 94.42 & 98.43 & 85.47 & 90.68 & 87.38 & 80.83 & 53.12 & 60.00 & 27.22 & 72.85 \\
\quad +single-VLM GMM & 51.88 & 48.53 & 64.99 & 94.00 & 97.30 & 68.02 & 74.74 & 83.69 & 69.65 & 50.26 & 42.50 & 18.67 & 63.69 \\
\midrule
QuiltNet-B/16 & 39.69 & 15.98 & 49.18 & 45.49 & 79.53 & 63.37 & 28.38 & 76.25 & 50.16 & 63.92 & 35.83 & 18.71 & 47.21 \\
\quad +multi-VLM GMM & 57.65 & 62.31 & 70.73 & 92.98 & 94.37 & 84.88 & 87.84 & 85.07 & 76.64 & 51.57 & 60.83 & 25.20 & 70.84 \\
\quad +single-VLM GMM & 53.22 & 30.49 & 44.72 & 72.02 & 93.13 & 63.08 & 51.98 & 83.39 & 67.80 & 52.60 & 48.33 & 11.72 & 56.04 \\
\midrule
QuiltNet-B/32 & 29.71 & 38.70 & 60.30 & 78.40 & 87.90 & 41.57 & 51.95 & 75.86 & 48.52 & 58.31 & 36.67 & 19.60 & 52.29 \\
\quad +multi-VLM GMM & 58.54 & 64.97 & 77.73 & 95.02 & 98.50 & 80.23 & 91.98 & 88.61 & 79.71 & 54.20 & 59.17 & 28.57 & 73.10 \\
\quad +single-VLM GMM & 18.63 & 47.99 & 71.71 & 89.20 & 98.20 & 46.80 & 66.66 & 86.43 & 74.31 & 49.68 & 40.00 & 16.90 & 58.88 \\
\midrule
CONCH & 26.61 & 53.73 & 64.83 & 89.51 & 97.23 & \underline{77.03} & 61.31 & 82.20 & 70.44 & \textbf{69.95} & 64.17 & 30.34 & 65.61 \\
\quad +multi-VLM GMM & 59.20 & 77.04 & 88.01 & 97.71 & 99.93 & 85.47 & 93.02 & 90.89 & 86.81 & 55.89 & 75.00 & 34.93 & 78.66 \\
\quad +single-VLM GMM & 57.43 & 70.01 & 88.40 & 97.62 & 99.73 & 82.85 & 87.02 & 91.02 & 86.55 & 62.35 & 72.50 & 28.32 & 76.98 \\
\midrule
UniMed-CLIP & 38.58 & 24.43 & 57.00 & 78.22 & 91.07 & 60.76 & 45.46 & 66.88 & 66.31 & 50.77 & 49.17 & 31.31 & 55.00 \\
\quad +multi-VLM GMM & 55.43 & 66.67 & 75.90 & 95.69 & 99.23 & 83.72 & 91.63 & 88.28 & 82.42 & 55.81 & 65.83 & 30.97 & 74.30 \\
\quad +single-VLM GMM & 50.78 & 30.57 & 68.41 & 94.80 & 99.00 & 61.05 & 89.42 & 87.22 & 66.26 & 45.39 & 54.17 & 28.53 & 64.63 \\
\midrule
KEEP & 32.37 & \underline{71.27} & \underline{86.53} & \underline{94.07} & 93.90 & 62.50 & \underline{84.99} & \underline{83.49} & 62.18 & 65.68 & \underline{68.33} & 32.24 & \underline{69.80} \\
\quad +multi-VLM GMM & 63.86 & 80.91 & 91.21 & 98.11 & 100.00 & 80.81 & 96.06 & 89.96 & 89.62 & 59.05 & 79.17 & 39.19 & 80.66 \\
\quad +single-VLM GMM & 47.01 & 81.38 & 91.19 & 98.02 & 100.00 & 66.86 & 96.09 & 86.49 & 85.17 & 59.06 & 75.00 & 38.73 & 77.08 \\
\midrule
PMC-CLIP & 23.73 & 18.19 & 46.31 & 74.64 & 63.63 & 47.97 & 51.62 & 75.79 & 59.48 & 50.90 & 46.67 & 17.99 & 48.08 \\
\quad +multi-VLM GMM & 53.44 & 66.59 & 76.32 & 95.18 & 98.47 & 79.65 & 90.60 & 87.32 & 79.29 & 49.60 & 62.50 & 26.04 & 72.08 \\
\quad +single-VLM GMM & 35.03 & 33.14 & 41.28 & 94.96 & 98.37 & 47.67 & 83.87 & 83.19 & 54.56 & 37.59 & 50.00 & 17.99 & 56.47 \\
\midrule
PathGen-CLIP & 37.25 & 42.94 & 66.87 & 85.02 & 96.63 & 66.86 & 57.23 & 81.54 & 74.36 & 64.89 & 60.83 & 33.84 & 64.02 \\
\quad +multi-VLM GMM & 60.09 & 69.90 & 81.48 & 95.87 & 99.40 & 80.81 & 90.61 & 90.36 & 82.42 & 55.98 & 69.17 & 30.30 & 75.53 \\
\quad +single-VLM GMM & 46.12 & 64.35 & 85.72 & 95.82 & 99.33 & 75.29 & 84.14 & 88.51 & 78.87 & 56.43 & 70.83 & 31.10 & 73.04 \\
\midrule
MUSK & 37.47 & 59.58 & 72.39 & 92.93 & \underline{99.23} & 66.57 & 66.91 & 75.43 & \textbf{82.73} & \underline{66.00} & 59.17 & \textbf{39.36} & 68.15 \\
\quad +multi-VLM GMM & 58.98 & 73.43 & 82.57 & 97.38 & 99.97 & 84.59 & 91.99 & 90.22 & 86.81 & 57.46 & 71.67 & 33.29 & 77.36 \\
\quad +single-VLM GMM & 47.89 & 69.76 & 66.23 & 97.42 & 99.97 & 85.17 & 92.33 & 89.27 & 89.04 & 56.88 & 70.00 & 35.95 & 74.99 \\
\midrule
OSTB & \textbf{58.54} & \textbf{79.78} & \textbf{87.88} & \textbf{97.84} & \textbf{99.67} & \textbf{84.30} & \textbf{94.54} & \textbf{90.36} & \underline{81.57} & 56.31 & \textbf{73.33} & \underline{35.86} & \textbf{78.33} \\
\bottomrule\end{tabular}
\end{table*}

\noindent\textbf{Medical-pathology integration.} Table~\ref{tab:pathology_results} shows that OSTB reaches 78.33\% on average, outperforming the strongest original pathology VLM, KEEP, at 69.80\%. In the main deployment comparison, OSTB gives the best result on 9 of the 12 datasets, with particularly clear gains on SICAP-MIL, SkinCancer, NCT-CRC, and WSSS4LUAD. This is a challenging setting because several specialist VLMs show strong dataset-specific preferences, yet the final OSTB ensemble still improves the domain average over the original VLM pool by integrating complementary semantic and visual evidence. OSTB is not the best deployment row on PanNuke, PCam, and BreakHis. The largest negative gap appears on PCam, where CONCH is substantially stronger than OSTB, and BreakHis favors MUSK. These results indicate that OSTB is not an oracle substitute for every individual dataset, but it provides a robust label-free deployment rule when the best specialist model is unknown in advance.

\noindent\textbf{Why does OSTB not win on every dataset?} Across the three domains, OSTB gives the best average accuracy among the main deployment rows in Tables~\ref{tab:natural_results}--\ref{tab:pathology_results}, but it is not the best deployment row on every individual benchmark. The exceptions follow different patterns. On Aircraft, PE-Core-B/16 is clearly stronger than OSTB, suggesting that this dataset is highly compatible with a particular general-purpose VLM. On several other natural-image datasets, such as CIFAR-10, the gap between OSTB and the best individual model is small because the strongest candidates are already highly accurate. In medical pathology, PCam favors CONCH by a large margin, while BreakHis favors MUSK, indicating that some pathology datasets are especially well matched to a specific specialist VLM. Importantly, these best individual models are identified using test labels for analysis; they are oracle references rather than deployment-time choices. OSTB instead applies a fixed label-free rule to each benchmark. Its role in Tables~\ref{tab:natural_results}--\ref{tab:pathology_results} is to provide a robust multi-model predictor without knowing in advance which candidate will be the column-wise oracle best. When compact or single-backbone deployment is preferred, the ranking and Top-$k$ results in Table~\ref{tab:topk_deployment} provide the corresponding mechanism for selecting a small reliable subset before deployment.

\subsection{Extended Studies}

\noindent\textbf{Component Ablation of OSTB.} We further decompose OSTB into semantic fusion, visual GMM fusion, transport marginal matching, adaptive weighting, and transport-conditioned GMM refinement. \textit{Average} directly averages the zero-shot semantic posteriors from all candidate VLMs. \textit{Semantic Ensemble} keeps only the semantic branch and fuses VLM semantic posteriors using the learned semantic reliability weights. \textit{GMM Ensemble} keeps only the adapted visual GMM branch. \textit{OSTB w/o transport marginal} keeps the iterative fusion process but replaces the transport update with row-wise normalization, thereby removing the marginal matching constraint. \textit{OSTB w/o GMM refinement} initializes the GMMs and keeps them fixed while still running transport and adaptive-weight updates. \textit{OSTB w/o adaptive weights} keeps the transport and GMM updates while fixing the branch-model weights during optimization. Table~\ref{tab:component_ablation} reports the resulting domain-wise averages and overall accuracy.

\begin{table}[!t]
\caption{Component ablation of OSTB. Domain-wise and overall Top-1 accuracy (\%) are reported after removing or isolating major semantic, visual, transport, refinement, and weighting components.}
\label{tab:component_ablation}
\centering
\scriptsize
\setlength{\tabcolsep}{4.5pt}
\renewcommand{\arraystretch}{1.08}
\resizebox{\columnwidth}{!}{%
\begin{tabular}{lcccc}
\toprule
Method & Natural & Remote & Pathology & Overall \\
\midrule
Average & 80.85 & 79.01 & 68.35 & 76.17 \\
Semantic Ensemble & 81.68 & 79.80 & 70.64 & 77.48 \\
GMM Ensemble & 68.09 & 85.18 & 76.87 & 75.77 \\
OSTB w/o transport marginal & \underline{84.39} & 84.74 & 75.77 & 81.62 \\
OSTB w/o GMM refinement & 83.01 & 82.40 & 75.32 & 80.28 \\
OSTB w/o adaptive weights & 83.98 & \textbf{85.41} & \underline{77.55} & \underline{82.24} \\
OSTB & \textbf{84.68} & \underline{85.37} & \textbf{78.33} & \textbf{82.75} \\
\bottomrule
\end{tabular}%
}
\end{table}

The full OSTB achieves the best accuracy on natural images, medical pathology, and the overall average, and it remains essentially tied with the best remote-sensing variant. Compared with direct averaging, OSTB improves the overall score from 76.17\% to 82.75\%, showing that naive probability averaging cannot reliably handle target-domain shift. Semantic-only fusion remains weaker at 77.48\%, indicating that semantic reliability alone is insufficient to capture the target-conditioned visual structure needed for robust deployment. The GMM-only branch is competitive on remote sensing and medical pathology but drops substantially on natural images, suggesting that visual GMMs are most effective when anchored by the original semantic signal. Removing the transport marginal constraint reduces the overall score to 81.62\%, while disabling GMM refinement or adaptive weights gives 80.28\% and 82.24\%, respectively. These results show that marginal matching, transport-conditioned GMM refinement, and adaptive branch weighting each make a measurable contribution to the final system.

\begin{figure}[!t]
\centering
\includegraphics[width=\columnwidth]{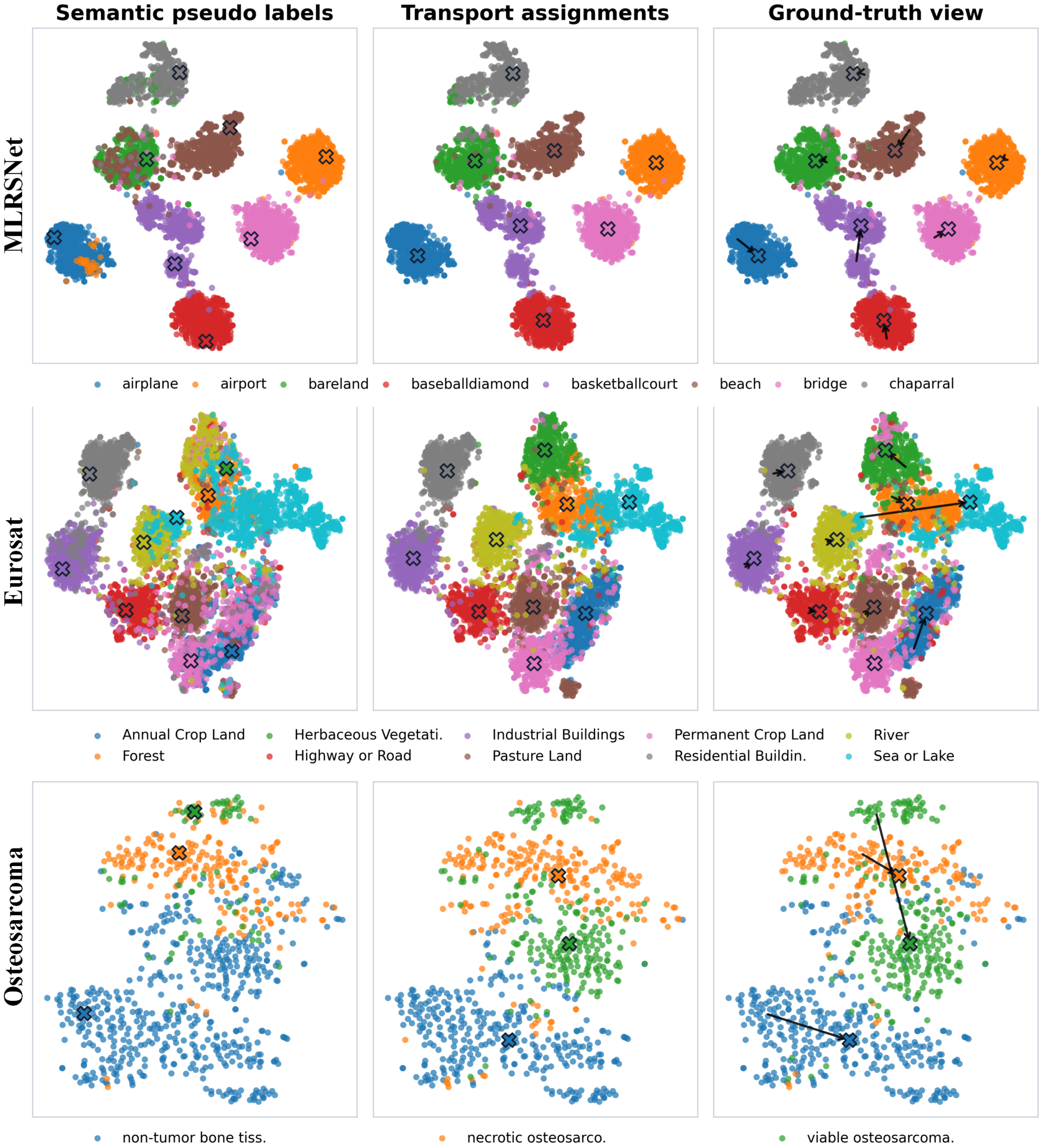}
\caption{Visualization of transport-induced GMM adaptation on representative target datasets from remote sensing, natural images, and medical pathology using CLIP ViT-B/16. Each row corresponds to one dataset, and the three columns show semantic pseudo labels, transport assignments, and the ground-truth view, respectively. Cross markers denote class-wise GMM centers. The arrows highlight that transport-conditioned adaptation moves the GMM centers from their initial positions toward class regions that are more visually coherent and semantically correct.}
\label{fig:transport_gmm_adaptation}
\end{figure}

\noindent\textbf{Transport-Induced GMM Adaptation.} To inspect how the visual branch is adapted, Fig.~\ref{fig:transport_gmm_adaptation} visualizes the GMM refinement process on one representative dataset from each domain. All three examples use the same CLIP ViT-B/16 backbone, so the differences mainly reflect target-domain structure rather than model-family changes. The left column shows that semantic pseudo labels can be noisy: visually adjacent samples may receive inconsistent labels, and some initial GMM centers are pulled away from dense class regions. After the transport update, the middle column exhibits more coherent sample-class assignments. The ground-truth view in the right column is used only for analysis, and it shows that adapted GMM centers move toward the corresponding visual clusters. This behavior is consistent across remote sensing, natural images, and pathology, indicating that the transport plan does not merely reweight predictions but also transfers shared target structure into the feature-space GMM branch. As a result, each Gaussian component becomes target-aware rather than remaining a fixed initialization inherited from noisy semantic pseudo labels.

\begin{figure*}[!t]
\centering
\captionsetup[subfloat]{font=footnotesize} 
\subfloat[Natural images]{%
\includegraphics[width=0.32\textwidth]{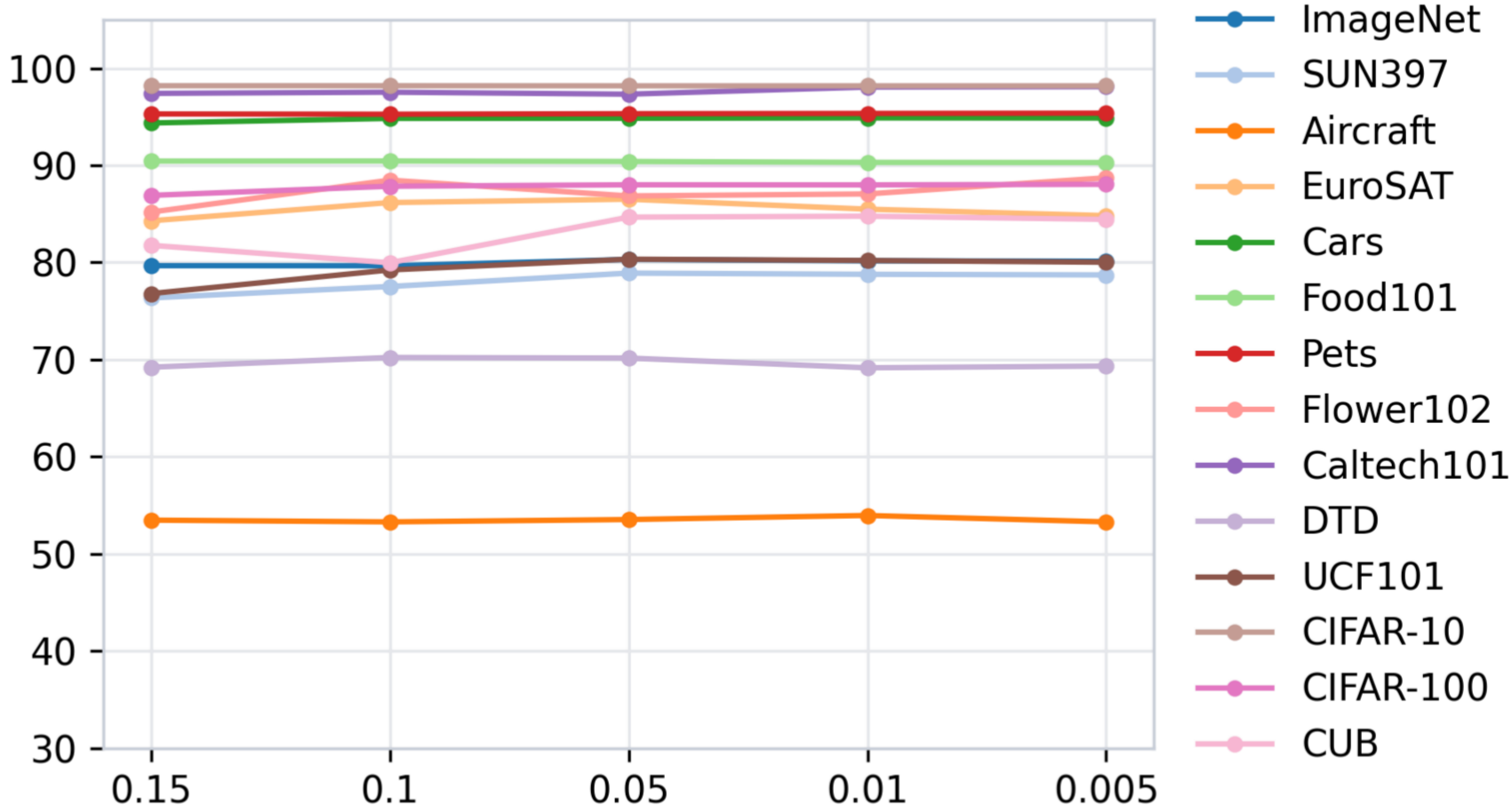}}
\hfill
\subfloat[Remote sensing]{%
\includegraphics[width=0.32\textwidth]{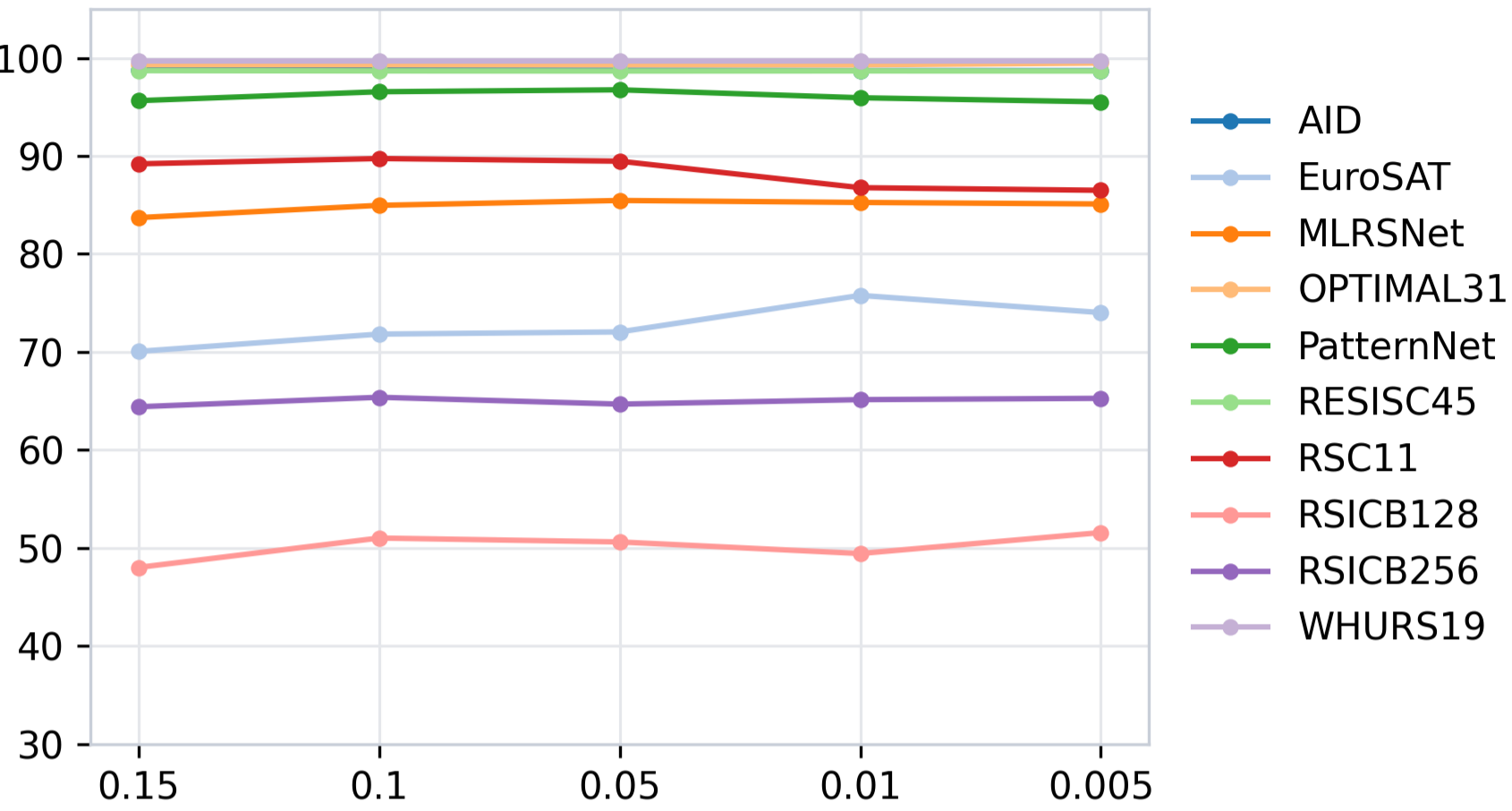}}
\hfill
\subfloat[Medical pathology]{%
\includegraphics[width=0.337\textwidth]{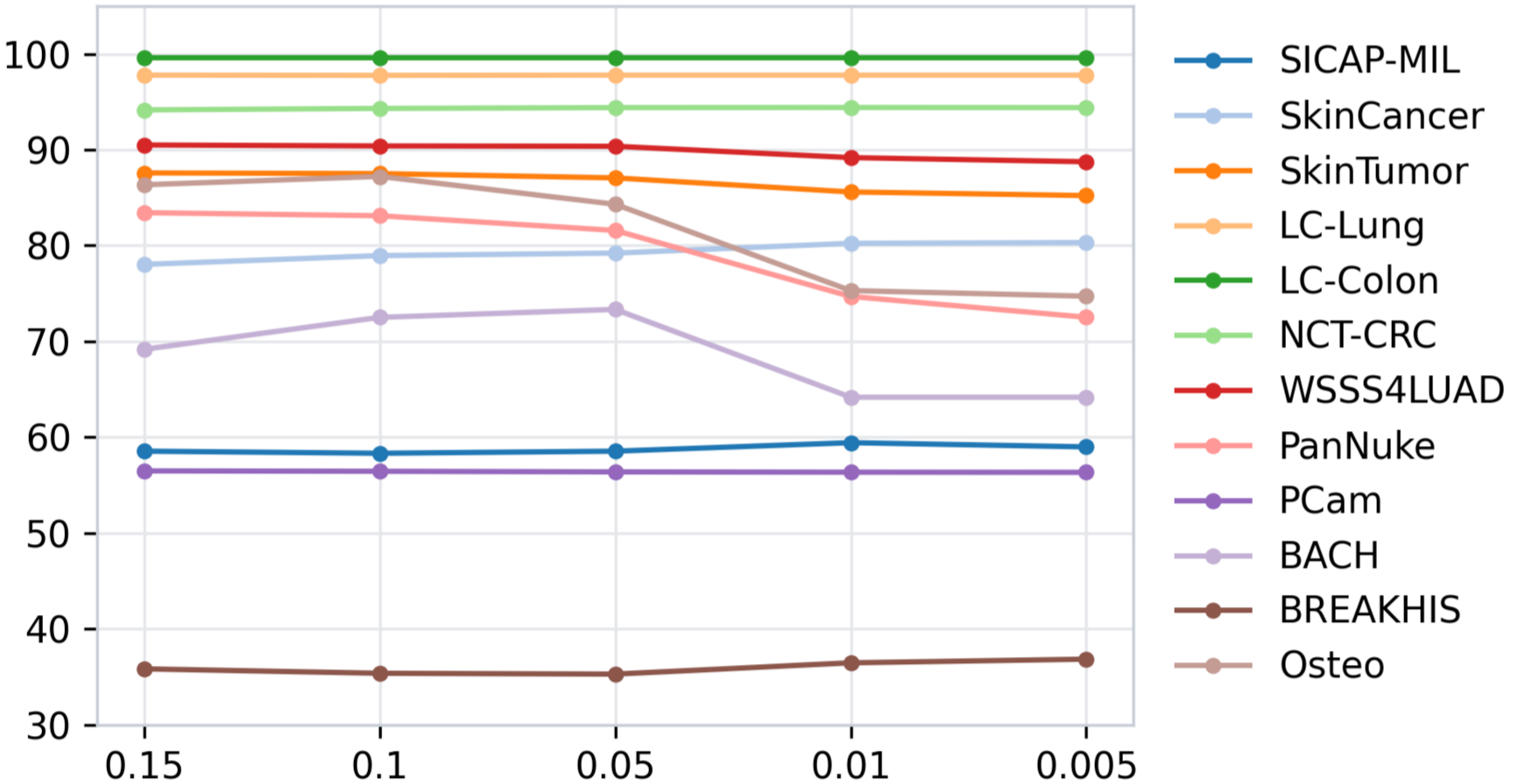}}
\caption{Parameter analysis of the entropic regularization coefficient $\epsilon$. The horizontal axis denotes the value of $\epsilon$, and the vertical axis reports top-1 accuracy (\%). Each curve corresponds to one dataset, with the three subfigures showing natural-image, remote-sensing, and medical-pathology benchmarks, respectively.}
\label{fig:eps_parameter_analysis}
\end{figure*}

\noindent\textbf{Unlabeled Adaptation Sample Size.} We next examine how much unlabeled target adaptation data is needed to estimate the transport plan, reliability weights, and transport-conditioned GMMs. For each benchmark, we randomly sample a fixed fraction of the unlabeled target adaptation split and keep the held-out target test split unchanged.

\begin{table}[!t]
\caption{Unlabeled target adaptation sample-size ablation. Domain-wise and overall Top-1 accuracy (\%) are reported when different fractions of the unlabeled adaptation split are used while the held-out test split remains unchanged.}
\label{tab:sample_size_ablation}
\centering
\footnotesize
\setlength{\tabcolsep}{6.0pt}
\renewcommand{\arraystretch}{1.08}
\begin{tabular}{lcccc}
\toprule
Target Adapt. Samples & Natural & Remote & Pathology & Overall \\
\midrule
10\% & 83.59 & 85.66 & 76.40 & 81.77 \\
25\% & 84.42 & 85.54 & 77.19 & 82.32 \\
50\% & 84.62 & \textbf{86.39} & 78.11 & \underline{82.94} \\
75\% & \textbf{84.86} & \underline{86.06} & \underline{78.14} & \textbf{82.95} \\
100\% & \underline{84.68} & 85.37 & \textbf{78.33} & 82.75 \\
\bottomrule
\end{tabular}
\end{table}

Table~\ref{tab:sample_size_ablation} reports the domain-wise and overall averages over all 36 benchmarks. Increasing the unlabeled subset from a very small ratio to a moderate ratio is helpful, but the gain from larger subsets is modest. The overall score increases from 81.77\% at 10\% of the target adaptation split to 82.94\% at 50\%, after which the curve plateaus, with 75\% and 100\% reaching 82.95\% and 82.75\%, respectively. Natural images and medical pathology follow the same broad pattern, while remote sensing peaks around 50\% but remains highly competitive over the larger subsets. The domain-wise curves are not perfectly monotonic, which is expected because changing the target adaptation subset size also changes class coverage and sample composition, especially on imbalanced or fine-grained benchmarks. Overall, OSTB is not strongly sample-hungry: moderate subsets already capture most of the benefit, and using all available target adaptation samples does not necessarily yield a higher average accuracy.

\begin{table}[!t]
\caption{Model-ranking quality under different unlabeled target adaptation sample sizes. Ranking metrics are computed against the oracle zero-shot model ranking on the held-out target test split.}
\label{tab:sample_size_ranking}
\centering
\scriptsize
\setlength{\tabcolsep}{4.0pt}
\renewcommand{\arraystretch}{1.08}
\resizebox{\columnwidth}{!}{%
\begin{tabular}{lcccccc}
\toprule
Adaptation Samples & $\rho\uparrow$ & $\tau\uparrow$ & Hit@1$\uparrow$ & Top-3$\uparrow$ & Regret$\downarrow$ & Sel.Acc$\uparrow$ \\
\midrule
10\% & 0.806 & \underline{0.682} & 0.639 & 0.917 & 1.239 & 77.04 \\
25\% & 0.804 & 0.679 & \textbf{0.722} & 0.917 & \textbf{1.114} & \textbf{77.16} \\
50\% & 0.806 & 0.681 & \underline{0.694} & 0.917 & \underline{1.129} & \underline{77.15} \\
75\% & \textbf{0.807} & \textbf{0.684} & \underline{0.694} & 0.917 & 1.312 & 76.96 \\
100\% & \underline{0.807} & 0.680 & 0.667 & 0.917 & 1.249 & 77.03 \\
\bottomrule
\end{tabular}
}
\end{table}

The ranking indicators in Table~\ref{tab:sample_size_ranking} show that the OSTB ranking score remains stable across different unlabeled target adaptation subset sizes. Spearman's $\rho$ stays between 0.804 and 0.807, Kendall's $\tau$ stays between 0.679 and 0.684, and Top-3 remains 0.917 for all tested ratios. The deployment-oriented metrics also vary mildly: Hit@1 ranges from 0.639 to 0.722, regret remains around 1.11--1.31 points, and Sel.Acc stays within 76.96--77.16\%. No single ratio dominates all metrics: the 75\% subset gives the strongest rank-correlation values, while the 25\% subset gives the best Hit@1, regret, and selected-model accuracy. Tables~\ref{tab:sample_size_ablation} and \ref{tab:sample_size_ranking} indicate that moderate unlabeled target adaptation subsets already provide enough target structure for reliable ranking and adaptation, while using all available target adaptation samples gives limited additional benefit.

\noindent\textbf{Candidate Pool Size.} We next examine how OSTB behaves under different candidate-pool sizes. We construct smaller pools in two ways. The first uses an oracle ordering, where candidate VLMs are sorted by their standalone accuracy on the target adaptation split using labels only for analysis. The second uses the OSTB ranking, where only the highest-ranked candidates are kept before rerunning the full adaptation-and-deployment procedure. Table~\ref{tab:pool_size_ablation} reports the corresponding domain-wise averages and overall accuracy.

\begin{table}[!t]
\caption{Candidate pool-size ablation of OSTB. Oracle-ranked and OSTB-ranked subsets are both rerun through the full adaptation-and-deployment procedure before reporting domain-wise and overall Top-1 accuracy (\%).}
\label{tab:pool_size_ablation}
\centering
\footnotesize
\setlength{\tabcolsep}{8.0pt}
\renewcommand{\arraystretch}{1.08}
\begin{tabular}{lcccc}
\toprule
Candidate Pool & Natural & Remote & Pathology & Overall \\
\midrule
\multicolumn{5}{c}{Oracle ranking} \\
Top-1 & 84.96 & 82.86 & 78.89 & 82.36 \\
Top-2 & 85.76 & 83.78 & 79.38 & 83.08 \\
Top-4 & \textbf{85.89} & 85.56 & \textbf{80.07} & \underline{83.86} \\
Top-6 & \underline{85.87} & \textbf{87.10} & \underline{80.01} & \textbf{84.26} \\
Top-8 & 85.46 & \underline{85.68} & 78.52 & 83.21 \\
\midrule
\multicolumn{5}{c}{OSTB ranking} \\
Top-1 & 84.62 & 79.95 & 76.62 & 80.66 \\
Top-2 & 85.59 & 83.38 & \textbf{78.39} & 82.57 \\
Top-4 & \textbf{85.83} & \textbf{86.24} & \underline{78.25} & \textbf{83.42} \\
Top-6 & \underline{85.62} & 85.45 & 77.31 & 82.80 \\
Top-8 & 85.51 & \underline{85.84} & 77.95 & \underline{83.08} \\
\midrule
All & 84.68 & 85.37 & 78.33 & 82.75 \\
\bottomrule
\end{tabular}
\end{table}

The oracle block provides an analysis-only reference for how performance changes when the strongest standalone models are retained first. Its overall accuracy increases from Top-1 to Top-6 and then drops at Top-8, indicating that a small group of strong candidates can provide useful complementarity, but adding more weaker models is not always beneficial. The OSTB-ranked block follows the same main trend without using target labels: performance improves rapidly from Top-1 to Top-4, drops at Top-6, and remains below the Top-4 peak at Top-8. Although the OSTB ranking is not identical to the oracle ordering, its compact subsets are competitive with the oracle subsets. For example, OSTB Top-4 reaches 83.42\% overall, close to the oracle Top-4 result of 83.86\%, and already surpasses using all candidates. This comparison shows that the learned ranking is useful not only as a reliability score, but also as a practical way to construct smaller deployment pools. Most of the benefit of multi-model deployment is retained by the highest-ranked few candidates, while using the full pool is not always optimal and can slightly reduce the overall average.

\noindent\textbf{Negative Model Injection.} We further test whether OSTB remains robust when the candidate pool is intentionally contaminated by clearly mismatched models. Starting from the clean in-domain candidate pool used in the main experiments, we inject pathology-specific VLMs into natural-image and remote-sensing tasks, and remote-sensing VLMs into medical-pathology tasks. In all cases, OSTB is rerun from scratch on the enlarged pool, so the reported accuracy reflects the full adaptation-and-deployment procedure rather than a post-hoc reweighting step. The resulting domain-wise and overall accuracies under the clean and injected candidate pools are reported in Table~\ref{tab:negative_injection}.

\begin{table}[!t]
\caption{Negative model injection ablation of OSTB. The clean candidate pools are compared with injected pools containing clearly mismatched cross-domain VLMs, with OSTB rerun from scratch in each setting.}
\label{tab:negative_injection}
\centering
\footnotesize
\setlength{\tabcolsep}{7.0pt}
\renewcommand{\arraystretch}{1.08}
\begin{tabular}{lcccc}
\toprule
Candidate Pool & Natural & Remote & Pathology & Overall \\
\midrule
Clean & 84.68 & 85.37 & 78.33 & 82.75 \\
Injected & 84.72 & 85.00 & 78.20 & 82.62 \\
$\Delta$ & +0.04 & -0.37 & -0.13 & -0.13 \\
\bottomrule
\end{tabular}
\end{table}

The injected distractors have only a limited effect on OSTB, although the three domains respond somewhat differently. Natural-image deployment is essentially unchanged and improves by 0.04 points on average. This tiny gain is more plausibly attributed to noise from the injected models than to any real contribution from pathology-specific knowledge, and it indicates that out-of-domain candidates do not seriously contaminate the original model pool. Remote-sensing and medical-pathology benchmarks show slight degradations, which is expected because these settings rely more strongly on domain-specific structure, so clearly mismatched cross-domain models can introduce mild interference during adaptation and integration. The drops are nevertheless small in absolute terms, especially for pathology. Overall, OSTB decreases by only 0.13 points under candidate-pool contamination, indicating that the clean in-domain models still dominate final decision.

\subsection{Parameter Analysis}

\noindent\textbf{Effect of Entropic Regularization.} We first analyze the sensitivity of OSTB to the entropic regularization coefficient $\epsilon$ in the transport update. Fig.~\ref{fig:eps_parameter_analysis} reports dataset-level results in the three application domains. Across the tested range from 0.15 to 0.005, most natural-image and remote-sensing datasets show only modest fluctuations, indicating that OSTB is not overly dependent on narrowly tuned regularization. A few benchmarks, especially in pathology, exhibit larger changes when visual clusters are less compact or class proportions make the marginal constraint more influential. Overall, the curves support using a single moderate value of $\epsilon$ across benchmarks rather than tuning it separately for every dataset.

\begin{figure}[!t]
\centering
\includegraphics[width=0.8\columnwidth]{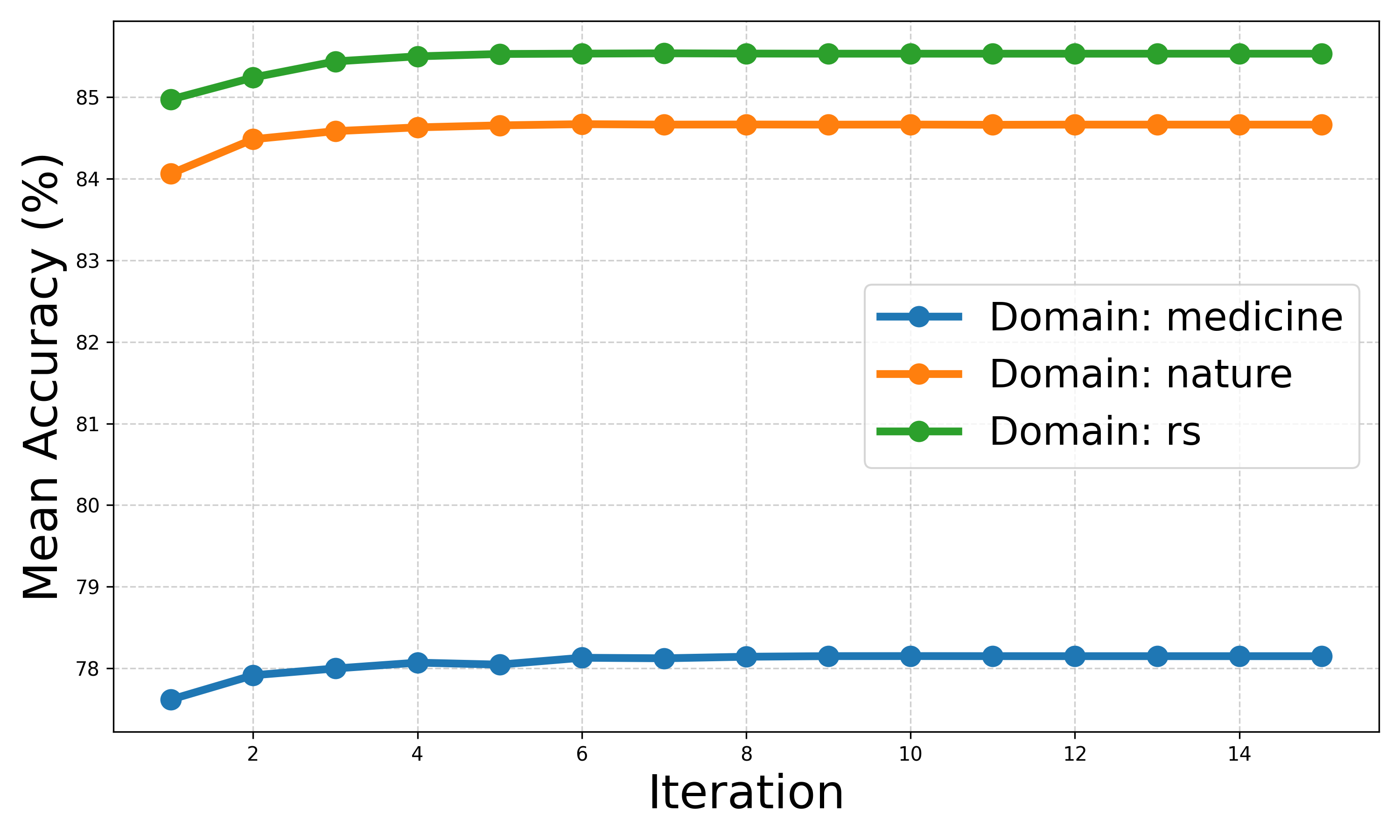}
\caption{Iteration-number analysis of OSTB. The horizontal axis denotes the number of alternating iterations, and the vertical axis reports the domain-wise mean Top-1 accuracy (\%).}
\label{fig:mm_iteration_analysis}
\end{figure}

\noindent\textbf{Effect of Iterations.} We further analyze the convergence behavior of OSTB by varying the number of alternating iterations from 1 to 15. As shown in Fig.~\ref{fig:mm_iteration_analysis}, all three domains improve rapidly in the first few iterations and then stabilize. Natural-image and remote-sensing benchmarks reach a plateau after several updates, while medical-pathology benchmarks also converge smoothly with only mild late-stage variation. This trend indicates that the transport-conditioned iterative update is effective but does not require excessive optimization, so a fixed small iteration budget is adequate for deployment.

\subsection{Computational Cost Analysis}

\begin{table}[!t]
	\caption{Runtime and memory analysis of OSTB from cached VLM features. The table reports adaptation/test sample counts, candidate-pool size, OSTB running time, peak GPU memory, and host RSS memory.}
	\label{tab:time_memory_analysis}
	\centering
	{\fontsize{6pt}{7pt}\selectfont
	\setlength{\tabcolsep}{2.2pt}
	\renewcommand{\arraystretch}{1.08}
		\begin{tabular}{@{}lccccccc@{}}
			\toprule
			Group & \#Data & \#Adapt & \#Test & \#VLMs & \shortstack{OSTB\\(s)} & \shortstack{GPU\\(GB)} & \shortstack{RSS\\(GB)} \\
			\midrule
			ImageNet & 1 & 1,281,167 & 50,000 & 14.0 & 2060.80 & 0.301 & 120.18 \\
			Natural w/o ImageNet & 13 & 15,087.8 & 8,422.3 & 14.0 & 7.49 & 0.145 & 3.90 \\
			Remote sensing & 10 & 19,152.6 & 8,208.6 & 9.0 & 7.24 & 0.092 & 2.94 \\
			Medical pathology & 12 & 38,925.8 & 7,711.8 & 12.0 & 14.43 & 0.111 & 8.62 \\
			\bottomrule
		\end{tabular}%
	}
\end{table}

Table~\ref{tab:time_memory_analysis} reports the computational overhead of OSTB after VLM features and probability outputs have been cached. 
This setting isolates the cost introduced by the proposed transport-based deployment module from the one-time forward passes of the candidate VLMs. 
It also matches the intended workflow: OSTB is used as a pre-deployment adaptation stage, where an unlabeled target adaptation split is first encoded by the candidate models and then used to estimate the shared transport plan, reliability weights, and transport-conditioned GMM classifiers. 
After this stage, transport optimization is no longer repeated for each held-out target test sample. 
During practical deployment, each held-out target test sample only needs to pass through the retained VLMs and the lightweight GMM classifiers for inference. 
When computational resources are extremely limited, the adapted GMM branch can also be deployed alone as a lightweight target-aware classifier.

The cost is modest on all non-ImageNet groups. 
Excluding ImageNet, OSTB takes 7.49s on natural-image benchmarks, 7.24s on remote-sensing benchmarks, and 14.43s on medical-pathology benchmarks on average, while peak GPU reserved memory remains below 0.15GB. 
This efficiency comes from the fact that the transport plan, GMM statistics, and test-time posteriors can be computed in chunks, so the GPU does not need to materialize large dense multi-model tensors at once. 
ImageNet is reported separately because it contains over 1.28 million unlabeled target adaptation samples and 14 candidate VLMs. 
Even at this scale, OSTB completes adaptation and evaluation in 2060.80s with only 0.301GB peak GPU memory. 
The main resource bottleneck is instead host memory: the ImageNet run reaches 120.18GB RSS because cached features from all candidate VLMs are resident in memory. 
These results show that OSTB adds a lightweight optimization layer on top of cached VLM outputs, and that its practical scalability is mainly governed by the size of the cached feature bank rather than by GPU memory. 
Ranking-guided Top-$k$ deployment can further reduce this overhead by retaining only a compact set of reliable VLMs before test-time inference.

\section{Conclusion}

In this paper, we presented OSTB, a self-adaptive optimal transport framework for selection-centered deployment with multiple VLMs. OSTB uses a shared transport structure to rank candidate models, adapt each model through transport-conditioned GMMs, and integrate multiple predictors in a reliability-aware manner. Thus, model selection, target-domain adaptation, and multi-model fusion are handled within a single unified framework. Experiments on natural-image, remote-sensing, and medical-pathology benchmarks demonstrate strong model ranking, effective single-model adaptation, and robust multi-model deployment compared with existing baselines. Future work will extend OSTB to broader multimodal foundation-model pools and explore stronger self-supervised target adaptation objectives and more scalable deployment strategies.

\bibliographystyle{IEEEtran}
\bibliography{ostb_refs}

@IEEEtranBSTCTL{IEEEtranBSTCTL:noDash,
  CTLdash_repeated_names = "no"
}

@inproceedings{radford2021learning,
  title={Learning Transferable Visual Models From Natural Language Supervision},
  author={Radford, Alec and Kim, Jong Wook and Hallacy, Chris and Ramesh, Aditya and Goh, Gabriel and Agarwal, Sandhini and Sastry, Girish and Askell, Amanda and Mishkin, Pamela and Clark, Jack and Krueger, Gretchen and Sutskever, Ilya},
  booktitle={Proceedings of the 38th International Conference on Machine Learning ({ICML})},
  series={Proceedings of Machine Learning Research},
  volume={139},
  pages={8748--8763},
  year={2021}
}

@inproceedings{cherti2023reproducible,
  title={Reproducible Scaling Laws for Contrastive Language-Image Learning},
  author={Cherti, Mehdi and Beaumont, Romain and Wightman, Ross and Wortsman, Mitchell and Ilharco, Gabriel and Gordon, Cade and Schuhmann, Christoph and Schmidt, Ludwig and Jitsev, Jenia},
  booktitle={Proceedings of the IEEE/CVF Conference on Computer Vision and Pattern Recognition ({CVPR})},
  pages={2818--2829},
  year={2023}
}

@article{sun2023evaclip,
  title={{EVA-CLIP}: Improved Training Techniques for {CLIP} at Scale},
  author={Sun, Quan and Fang, Yuxin and Wu, Ledell and Wang, Xinlong and Cao, Yue},
  journal={arXiv preprint arXiv:2303.15389},
  year={2023}
}

@inproceedings{xu2024demystifying,
  title={Demystifying {CLIP} Data},
  author={Xu, Hu and Xie, Saining and Tan, Xiaoqing and Huang, Po-Yao and Howes, Russell and Sharma, Vasu and Li, Shang-Wen and Ghosh, Gargi and Zettlemoyer, Luke and Feichtenhofer, Christoph},
  booktitle={The Twelfth International Conference on Learning Representations ({ICLR})},
  year={2024}
}

@article{siglip,
  title={{SigLIP} 2: Multilingual Vision-Language Encoders with Improved Semantic Understanding, Localization, and Dense Features},
  author={Tschannen, Michael and Gritsenko, Alexey and Wang, Xiao and Naeem, Muhammad Ferjad and Alabdulmohsin, Ibrahim and Parthasarathy, Nikhil and Evans, Talfan and Beyer, Lucas and Xia, Ye and Mustafa, Basil and H{\'e}naff, Olivier and Harmsen, Jeremiah and Steiner, Andreas and Zhai, Xiaohua},
  journal={arXiv preprint arXiv:2502.14786},
  year={2025}
}

@article{MM,
  title={A Tutorial on {MM} Algorithms},
  author={Hunter, David R. and Lange, Kenneth},
  journal={The American Statistician},
  volume={58},
  number={1},
  pages={30--37},
  year={2004}
}

@article{conchlu2024visual,
  title={A Visual-Language Foundation Model for Computational Pathology},
  author={Lu, Ming Y. and Chen, Bowen and Williamson, Drew F. K. and Chen, Richard J. and Liang, Ivy and Ding, Tong and Jaume, Guillaume and Odintsov, Igor and Le, Long Phi and Gerber, Georg and Parwani, Anil V. and Zhang, Andrew and Mahmood, Faisal},
  journal={Nature Medicine},
  volume={30},
  number={3},
  pages={863--874},
  year={2024},
  doi={10.1038/s41591-024-02856-4}
}

@article{pliphuang2023visual,
  title={A Visual-Language Foundation Model for Pathology Image Analysis Using Medical Twitter},
  author={Huang, Zhi and Bianchi, Federico and Yuksekgonul, Mert and Montine, Thomas J. and Zou, James},
  journal={Nature Medicine},
  volume={29},
  number={9},
  pages={2307--2316},
  year={2023},
  doi={10.1038/s41591-023-02504-3}
}

@article{remoteclip,
  title={{RemoteCLIP}: A Vision Language Foundation Model for Remote Sensing},
  author={Liu, Fan and Chen, Delong and Guan, Zhangqingyun and Zhou, Xiaocong and Zhu, Jiale and Ye, Qiaolin and Fu, Liyong and Zhou, Jun},
  journal={IEEE Transactions on Geoscience and Remote Sensing},
  volume={62},
  pages={1--16},
  year={2024},
  doi={10.1109/TGRS.2024.3390838}
}

@article{georsclip,
  title={{RS5M} and {GeoRSCLIP}: A Large Scale Vision-Language Dataset and A Large Vision-Language Model for Remote Sensing},
  author={Zhang, Zilun and Zhao, Tiancheng and Guo, Yulong and Yin, Jianwei},
  journal={IEEE Transactions on Geoscience and Remote Sensing},
  volume={62},
  pages={1--23},
  year={2024},
  doi={10.1109/TGRS.2024.3449154}
}

@article{muskxiang2025vision,
  title={A Vision--Language Foundation Model for Precision Oncology},
  author={Xiang, Jinxi and Wang, Xiyue and Zhang, Xiaoming and Xi, Yinghua and Eweje, Feyisope and Chen, Yijiang and Li, Yuchen and Bergstrom, Colin and Gopaulchan, Matthew and Kim, Ted and Yu, Kun-Hsing and Willens, Sierra and Olguin, Francesca Maria and Nirschl, Jeffrey J. and Neal, Joel and Diehn, Maximilian and Yang, Sen and Li, Ruijiang},
  journal={Nature},
  volume={638},
  pages={769--778},
  year={2025},
  doi={10.1038/s41586-024-08378-w}
}

@article{zhou2024keep,
  title={Knowledge-enhanced Pretraining for Vision-language Pathology Foundation Model on Cancer Diagnosis},
  author={Zhou, Xiao and Sun, Luoyi and He, Dexuan and Guan, Wenbin and Wang, Ge and Wang, Ruifen and Wang, Lifeng and Yuan, Xiaojun and Sun, Xin and Zhang, Ya and Sun, Kun and Wang, Yanfeng and Xie, Weidi},
  journal={Cancer Cell},
  volume={44},
  number={4},
  pages={777--791},
  year={2026}
}

@inproceedings{ikezogwo2023quilt,
  title={Quilt-1M: One Million Image-Text Pairs for Histopathology},
  author={Ikezogwo, Wisdom Oluchi and Seyfioglu, Mehmet Saygin and Ghezloo, Fatemeh and Geva, Dylan Stefan Chan and Mohammed, Fatwir Sheikh and Anand, Pavan Kumar and Krishna, Ranjay and Shapiro, Linda},
  booktitle={Advances in Neural Information Processing Systems},
  volume={36},
  pages={37995--38017},
  year={2023}
}

@article{biomedclip,
  title={A Multimodal Biomedical Foundation Model Trained from Fifteen Million Image-Text Pairs},
  author={Zhang, Sheng and Xu, Yanbo and Usuyama, Naoto and Xu, Hanwen and Bagga, Jaspreet and Tinn, Robert and Preston, Sam and Rao, Rajesh and Wei, Mu and Valluri, Naveen and Wong, Cliff and Tupini, Andrea and Wang, Yu and Mazzola, Matt and Shukla, Swadheen and Liden, Lars and Gao, Jianfeng and Crabtree, Angela and Piening, Brian and Bifulco, Carlo and Lungren, Matthew P. and Naumann, Tristan and Wang, Sheng and Poon, Hoifung},
  journal={{NEJM AI}},
  volume={2},
  number={1},
  pages={AIoa2400640},
  year={2025},
  doi={10.1056/AIoa2400640}
}

@article{khattak2024unimed,
  title={{UniMed-CLIP}: Towards a Unified Image-Text Pretraining Paradigm for Diverse Medical Imaging Modalities},
  author={Khattak, Muhammad Uzair and Kunhimon, Shahina and Naseer, Muzammal and Khan, Salman and Khan, Fahad Shahbaz},
  journal={arXiv preprint arXiv:2412.10372},
  year={2024}
}

@inproceedings{skyscript,
  title={{SkyScript}: A Large and Semantically Diverse Vision-Language Dataset for Remote Sensing},
  author={Wang, Zhecheng and Prabha, Rajanie and Huang, Tianyuan and Wu, Jiajun and Rajagopal, Ram},
  booktitle={Proceedings of the AAAI Conference on Artificial Intelligence ({AAAI})},
  volume={38},
  number={6},
  pages={5805--5813},
  year={2024}
}

@inproceedings{TPT,
  title={Test-Time Prompt Tuning for Zero-Shot Generalization in Vision-Language Models},
  author={Shu, Manli and Nie, Weili and Huang, De-An and Yu, Zhiding and Goldstein, Tom and Anandkumar, Anima and Xiao, Chaowei},
  booktitle={Advances in Neural Information Processing Systems ({NeurIPS})},
  volume={35},
  pages={14274--14289},
  year={2022}
}

@article{CoOp,
  title={Learning to Prompt for Vision-Language Models},
  author={Zhou, Kaiyang and Yang, Jingkang and Loy, Chen Change and Liu, Ziwei},
  journal={International Journal of Computer Vision},
  volume={130},
  number={9},
  pages={2337--2348},
  year={2022},
  doi={10.1007/s11263-022-01653-1}
}

@inproceedings{CoCoOp,
  title={Conditional Prompt Learning for Vision-Language Models},
  author={Zhou, Kaiyang and Yang, Jingkang and Loy, Chen Change and Liu, Ziwei},
  booktitle={Proceedings of the IEEE/CVF Conference on Computer Vision and Pattern Recognition ({CVPR})},
  pages={16816--16825},
  year={2022}
}

@inproceedings{lafter,
  title={{LaFTer}: Label-Free Tuning of Zero-Shot Classifier using Language and Unlabeled Image Collections},
  author={Mirza, Muhammad Jehanzeb and Karlinsky, Leonid and Lin, Wei and Kozinski, Mateusz and Possegger, Horst and Feris, Rogerio and Bischof, Horst},
  booktitle={Advances in Neural Information Processing Systems ({NeurIPS})},
  volume={36},
  pages={5765--5777},
  year={2023}
}

@inproceedings{transductive-3,
  title={Transductive Zero-Shot and Few-Shot {CLIP}},
  author={Martin, S{\'e}gol{\`e}ne and Huang, Yunshi and Shakeri, Fereshteh and Pesquet, Jean-Christophe and Ben Ayed, Ismail},
  booktitle={Proceedings of the IEEE/CVF Conference on Computer Vision and Pattern Recognition ({CVPR})},
  pages={28816--28826},
  year={2024}
}

@inproceedings{ZERO,
  title={Frustratingly Easy Test-Time Adaptation of Vision-Language Models},
  author={Farina, Matteo and Franchi, Gianni and Iacca, Giovanni and Mancini, Massimiliano and Ricci, Elisa},
  booktitle={Advances in Neural Information Processing Systems ({NeurIPS})},
  volume={37},
  pages={129062--129093},
  year={2024}
}

@inproceedings{karmanov2024efficient,
  title={Efficient Test-Time Adaptation of Vision-Language Models},
  author={Karmanov, Adilbek and Guan, Dayan and Lu, Shijian and El Saddik, Abdulmotaleb and Xing, Eric},
  booktitle={Proceedings of the IEEE/CVF Conference on Computer Vision and Pattern Recognition ({CVPR})},
  pages={14162--14171},
  year={2024}
}

@inproceedings{huang2025cosmic,
  title={{COSMIC}: Clique-Oriented Semantic Multi-Space Integration for Robust {CLIP} Test-Time Adaptation},
  author={Huang, Fanding and Jiang, Jingyan and Jiang, Qinting and Li, Hebei and Khan, Faisal Nadeem and Wang, Zhi},
  booktitle={Proceedings of the IEEE/CVF Conference on Computer Vision and Pattern Recognition ({CVPR})},
  pages={9772--9781},
  year={2025}
}

@article{oquab2024dinov2,
  title={{DINOv2}: Learning Robust Visual Features without Supervision},
  author={Oquab, Maxime and Darcet, Timoth{\'e}e and Moutakanni, Th{\'e}o and Vo, Huy V. and Szafraniec, Marc and Khalidov, Vasil and Fernandez, Pierre and Haziza, Daniel and Massa, Francisco and El-Nouby, Alaaeldin and Assran, Mahmoud and Ballas, Nicolas and Galuba, Wojciech and Howes, Russell and Huang, Po-Yao and Li, Shang-Wen and Misra, Ishan and Rabbat, Michael and Sharma, Vasu and Synnaeve, Gabriel and Xu, Hu and Jegou, Herve and Mairal, Julien and Labatut, Patrick and Joulin, Armand and Bojanowski, Piotr},
  journal={Transactions on Machine Learning Research},
  year={2024}
}

@inproceedings{DPE,
  title={Dual Prototype Evolving for Test-Time Generalization of Vision-Language Models},
  author={Zhang, Ce and Stepputtis, Simon and Sycara, Katia and Xie, Yaqi},
  booktitle={Advances in Neural Information Processing Systems ({NeurIPS})},
  volume={37},
  pages={32111--32136},
  year={2024}
}

@inproceedings{zhang2024dual,
  title={Dual Memory Networks: A Versatile Adaptation Approach for Vision-Language Models},
  author={Zhang, Yabin and Zhu, Wenjie and Tang, Hui and Ma, Zhiyuan and Zhou, Kaiyang and Zhang, Lei},
  booktitle={Proceedings of the IEEE/CVF Conference on Computer Vision and Pattern Recognition ({CVPR})},
  pages={28718--28728},
  year={2024}
}

@article{OT,
  title={Computational Optimal Transport: With Applications to Data Science},
  author={Peyr{\'e}, Gabriel and Cuturi, Marco},
  journal={Foundations and Trends{\textregistered} in Machine Learning},
  volume={11},
  number={5--6},
  pages={355--607},
  year={2019}
}

@inproceedings{chang2022unified,
  title={Unified Optimal Transport Framework for Universal Domain Adaptation},
  author={Chang, Wanxing and Shi, Ye and Tuan, Hoang Duong and Wang, Jingya},
  booktitle={Advances in Neural Information Processing Systems ({NeurIPS})},
  volume={35},
  pages={29512--29524},
  year={2022}
}

@inproceedings{chenplot,
  title={{PLOT}: Prompt Learning with Optimal Transport for Vision-Language Models},
  author={Chen, Guangyi and Yao, Weiran and Song, Xiangchen and Li, Xinyue and Rao, Yongming and Zhang, Kun},
  booktitle={The Eleventh International Conference on Learning Representations ({ICLR})},
  year={2023}
}

@inproceedings{tan2025recover,
  title={Recover and Match: Open-Vocabulary Multi-Label Recognition through Knowledge-Constrained Optimal Transport},
  author={Tan, Hao and Tan, Zichang and Li, Jun and Liu, Ajian and Wan, Jun and Lei, Zhen},
  booktitle={Proceedings of the IEEE/CVF Conference on Computer Vision and Pattern Recognition ({CVPR})},
  pages={4650--4660},
  year={2025}
}

@inproceedings{awt,
  title={{AWT}: Transferring Vision-Language Models via Augmentation, Weighting, and Transportation},
  author={Zhu, Yuhan and Ji, Yuyang and Zhao, Zhiyu and Wu, Gangshan and Wang, Limin},
  booktitle={Advances in Neural Information Processing Systems ({NeurIPS})},
  volume={37},
  pages={25561--25591},
  year={2024}
}

@inproceedings{lakshminarayanan2017simple,
  title={Simple and Scalable Predictive Uncertainty Estimation Using Deep Ensembles},
  author={Lakshminarayanan, Balaji and Pritzel, Alexander and Blundell, Charles},
  booktitle={Advances in Neural Information Processing Systems},
  volume={30},
  year={2017}
}

@inproceedings{wortsman2022model,
  title={Model Soups: Averaging Weights of Multiple Fine-Tuned Models Improves Accuracy Without Increasing Inference Time},
  author={Wortsman, Mitchell and Ilharco, Gabriel and Gadre, Samir Yitzhak and Roelofs, Rebecca and Gontijo Lopes, Raphael and Morcos, Ari S. and Namkoong, Hongseok and Farhadi, Ali and Carmon, Yair and Kornblith, Simon and Schmidt, Ludwig},
  booktitle={Proceedings of the 39th International Conference on Machine Learning ({ICML})},
  volume={162},
  year={2022}
}

@inproceedings{hu2026sota,
  title={SOTA: Self-adaptive Optimal Transport for Zero-Shot Classification with Multiple Foundation Models},
  author={Hu, Zhanxuan and Xu, Qiyu and Duan, Yu and Tai, Yonghang and Li, Huafeng},
  booktitle={Proceedings of the IEEE/CVF Conference on Computer Vision and Pattern Recognition ({CVPR})},
  pages={26624--26634},
  year={2026},
  month={June}
}

@inproceedings{imagenet,
  title={{ImageNet}: A Large-Scale Hierarchical Image Database},
  author={Deng, Jia and Dong, Wei and Socher, Richard and Li, Li-Jia and Li, Kai and Fei-Fei, Li},
  booktitle={Proceedings of the IEEE Conference on Computer Vision and Pattern Recognition ({CVPR})},
  pages={248--255},
  year={2009}
}

@inproceedings{sun397,
  title={{SUN} Database: Large-Scale Scene Recognition from Abbey to Zoo},
  author={Xiao, Jianxiong and Hays, James and Ehinger, Krista A. and Oliva, Aude and Torralba, Antonio},
  booktitle={Proceedings of the IEEE Conference on Computer Vision and Pattern Recognition ({CVPR})},
  pages={3485--3492},
  year={2010}
}

@article{aircraft,
  title={Fine-Grained Visual Classification of Aircraft},
  author={Maji, Subhransu and Rahtu, Esa and Kannala, Juho and Blaschko, Matthew and Vedaldi, Andrea},
  journal={arXiv preprint arXiv:1306.5151},
  year={2013}
}

@article{eurosat,
  title={{EuroSAT}: A Novel Dataset and Deep Learning Benchmark for Land Use and Land Cover Classification},
  author={Helber, Patrick and Bischke, Benjamin and Dengel, Andreas and Borth, Damian},
  journal={IEEE Journal of Selected Topics in Applied Earth Observations and Remote Sensing},
  volume={12},
  number={7},
  pages={2217--2226},
  year={2019}
}

@inproceedings{scars,
  title={{3D} Object Representations for Fine-Grained Categorization},
  author={Krause, Jonathan and Stark, Michael and Deng, Jia and Fei-Fei, Li},
  booktitle={Proceedings of the IEEE/CVF International Conference on Computer Vision Workshops ({ICCVW})},
  pages={554--561},
  year={2013}
}

@inproceedings{food101,
  title={{Food-101}: Mining Discriminative Components with Random Forests},
  author={Bossard, Lukas and Guillaumin, Matthieu and Van Gool, Luc},
  booktitle={Computer Vision -- {ECCV} 2014},
  pages={446--461},
  publisher={Springer},
  year={2014}
}

@inproceedings{pets,
  title={Cats and Dogs},
  author={Parkhi, Omkar M. and Vedaldi, Andrea and Zisserman, Andrew and Jawahar, C. V.},
  booktitle={Proceedings of the IEEE Conference on Computer Vision and Pattern Recognition ({CVPR})},
  pages={3498--3505},
  year={2012}
}

@inproceedings{flowers,
  title={Automated Flower Classification over a Large Number of Classes},
  author={Nilsback, Maria-Elena and Zisserman, Andrew},
  booktitle={Proceedings of the Sixth Indian Conference on Computer Vision, Graphics \& Image Processing},
  pages={722--729},
  year={2008}
}

@inproceedings{caltech101,
  title={Learning Generative Visual Models from Few Training Examples: An Incremental Bayesian Approach Tested on 101 Object Categories},
  author={Fei-Fei, Li and Fergus, Rob and Perona, Pietro},
  booktitle={Proceedings of the 2004 IEEE Computer Society Conference on Computer Vision and Pattern Recognition Workshops},
  pages={178},
  year={2004}
}

@inproceedings{dtd,
  title={Describing Textures in the Wild},
  author={Cimpoi, Mircea and Maji, Subhransu and Kokkinos, Iasonas and Mohamed, Sammy and Vedaldi, Andrea},
  booktitle={Proceedings of the IEEE Conference on Computer Vision and Pattern Recognition ({CVPR})},
  pages={3606--3613},
  year={2014}
}

@article{ucf101,
  title={{UCF101}: A Dataset of 101 Human Actions Classes From Videos in the Wild},
  author={Soomro, Khurram and Zamir, Amir Roshan and Shah, Mubarak},
  journal={arXiv preprint arXiv:1212.0402},
  year={2012}
}

@techreport{cifar,
  title={Learning Multiple Layers of Features from Tiny Images},
  author={Krizhevsky, Alex},
  institution={University of Toronto},
  year={2009}
}

@inproceedings{cub,
  title={The Caltech-{UCSD} Birds-200-2011 Dataset},
  author={Wah, Catherine and Branson, Steve and Welinder, Peter and Perona, Pietro and Belongie, Serge},
  booktitle={California Institute of Technology Technical Report {CNS-TR-2011-001}},
  year={2011}
}

@article{sicap,
  title={Proportion Constrained Weakly Supervised Histopathology Image Classification},
  author={Silva-Rodr{\'\i}guez, Julio and Schmidt, Arne and Sales, Mar{\'\i}a A. and Molina, Rafael and Naranjo, Valery},
  journal={Computers in Biology and Medicine},
  volume={147},
  pages={105714},
  year={2022},
  doi={10.1016/j.compbiomed.2022.105714}
}

@inproceedings{pcam,
  title={Rotation Equivariant {CNN}s for Digital Pathology},
  author={Veeling, Bastiaan S. and Linmans, Jasper and Winkens, Jim and Cohen, Taco and Welling, Max},
  booktitle={Medical Image Computing and Computer Assisted Intervention -- MICCAI 2018},
  pages={210--218},
  publisher={Springer},
  year={2018},
  doi={10.1007/978-3-030-00934-2_24}
}

@misc{osteosarcoma,
  title={Osteosarcoma Data from {UT} Southwestern/{UT} Dallas for Viable and Necrotic Tumor Assessment},
  author={Leavey, Patrick and Sengupta, Anita and Rakheja, Dinesh and Daescu, Ovidiu and Arunachalam, Harish Babu and Mishra, Rashika},
  howpublished={The Cancer Imaging Archive},
  year={2019},
  doi={10.7937/TCIA.2019.BVHJHDAS}
}

@article{bach,
  title={{BACH}: Grand Challenge on Breast Cancer Histology Images},
  author={Aresta, Guilherme and Ara{\'u}jo, Teresa and Kwok, Scotty and Chennamsetty, Sai Saketh and Safwan, Mohammed and Alex, Varghese and Marami, Bahram and Prastawa, Marcel and Chan, Monica and Donovan, Michael and others},
  journal={Medical Image Analysis},
  volume={56},
  pages={122--139},
  year={2019},
  doi={10.1016/j.media.2019.05.010}
}

@article{breakhis,
  title={A Dataset for Breast Cancer Histopathological Image Classification},
  author={Spanhol, Fabio A. and Oliveira, Luiz S. and Petitjean, Caroline and Heutte, Laurent},
  journal={IEEE Transactions on Biomedical Engineering},
  volume={63},
  number={7},
  pages={1455--1462},
  year={2016},
  doi={10.1109/TBME.2015.2496264}
}

@article{skincancer,
  title={Deep Learning for the Detection of Anatomical Tissue Structures and Neoplasms of the Skin on Scanned Histopathological Tissue Sections},
  author={Kriegsmann, Katharina and L{\"o}bers, Frithjof and Zgorzelski, Christiane and Kriegsmann, Joerg and Janssen, Charlotte and Meli{\ss}, Rolf Rudinger and Muley, Thomas and Sack, Ulrich and Steinbuss, Georg and Kriegsmann, Mark},
  journal={Frontiers in Oncology},
  volume={12},
  pages={1022967},
  year={2022},
  doi={10.3389/fonc.2022.1022967}
}

@article{lung,
  title={Lung and Colon Cancer Histopathological Image Dataset ({LC25000})},
  author={Borkowski, Andrew A. and Bui, Marilyn M. and Thomas, L. Brannon and Wilson, Catherine P. and DeLand, Lauren A. and Mastorides, Stephen M.},
  journal={arXiv preprint arXiv:1912.12142},
  year={2019}
}

@article{nct,
  title={100,000 Histological Images of Human Colorectal Cancer and Healthy Tissue},
  author={Kather, Jakob Nikolas and Halama, Niels and Marx, Alexander},
  journal={Zenodo},
  year={2018},
  doi={10.5281/zenodo.1214456}
}

@article{luad,
  title={Multi-layer Pseudo-supervision for Histopathology Tissue Semantic Segmentation Using Patch-level Classification Labels},
  author={Han, Chu and Lin, Jiatai and Mai, Jinhai and Wang, Yi and Zhang, Qingling and Zhao, Bingchao and Chen, Xin and Pan, Xipeng and Shi, Zhenwei and Xu, Zeyan and others},
  journal={Medical Image Analysis},
  volume={80},
  pages={102487},
  year={2022},
  doi={10.1016/j.media.2022.102487}
}

@article{gamper2020pannuke,
  title={{PanNuke} Dataset Extension, Insights and Baselines},
  author={Gamper, Jevgenij and Alemi Koohbanani, Navid and Benes, Ksenija and Graham, Simon and Jahanifar, Mostafa and Khurram, Syed Ali and Azam, Ayesha and Hewitt, Katherine and Rajpoot, Nasir},
  journal={arXiv preprint arXiv:2003.10778},
  year={2020}
}

@article{AID,
  title={{AID}: A Benchmark Data Set for Performance Evaluation of Aerial Scene Classification},
  author={Xia, Gui-Song and Hu, Jingwen and Hu, Fan and Shi, Baoguang and Bai, Xiang and Zhong, Yanfei and Zhang, Liangpei and Lu, Xiaoqiang},
  journal={IEEE Transactions on Geoscience and Remote Sensing},
  volume={55},
  number={7},
  pages={3965--3981},
  year={2017}
}

@article{MLRSNet,
  title={{MLRSNet}: A Multi-Label High Spatial Resolution Remote Sensing Dataset for Semantic Scene Understanding},
  author={Qi, Xiaoman and Zhu, Panpan and Wang, Yuebin and Zhang, Liqiang and Peng, Junhuan and Wu, Mengfan and Chen, Jialong and Zhao, Xudong and Zang, Ning and Mathiopoulos, P. Takis},
  journal={ISPRS Journal of Photogrammetry and Remote Sensing},
  volume={169},
  pages={337--350},
  year={2020}
}

@article{OPTIMAL,
  title={Scene Classification With Recurrent Attention of {VHR} Remote Sensing Images},
  author={Wang, Qi and Liu, Shaoteng and Chanussot, Jocelyn and Li, Xuelong},
  journal={IEEE Transactions on Geoscience and Remote Sensing},
  volume={57},
  number={2},
  pages={1155--1167},
  year={2019}
}

@article{PatternNet,
  title={{PatternNet}: A Benchmark Dataset for Performance Evaluation of Remote Sensing Image Retrieval},
  author={Zhou, Weixun and Newsam, Shawn and Li, Congmin and Shao, Zhenfeng},
  journal={ISPRS Journal of Photogrammetry and Remote Sensing},
  volume={145},
  pages={197--209},
  year={2018}
}

@article{RESISC45,
  title={Remote Sensing Image Scene Classification: Benchmark and State of the Art},
  author={Cheng, Gong and Han, Junwei and Lu, Xiaoqiang},
  journal={Proceedings of the IEEE},
  volume={105},
  number={10},
  pages={1865--1883},
  year={2017}
}

@article{RSC11,
  title={Feature Significance-Based Multi-Bag-of-Visual-Words Model for Remote Sensing Image Scene Classification},
  author={Zhao, Lijun and Tang, Ping and Huo, Lianzhi},
  journal={Journal of Applied Remote Sensing},
  volume={10},
  number={3},
  pages={035004},
  year={2016}
}

@article{RSICB,
  title={{RSI-CB}: A Large-Scale Remote Sensing Image Classification Benchmark Using Crowdsourced Data},
  author={Li, Haifeng and Dou, Xin and Tao, Chao and Wu, Zhixiang and Chen, Jie and Peng, Jian and Deng, Min and Zhao, Ling},
  journal={Sensors},
  volume={20},
  number={6},
  pages={1594},
  year={2020}
}

@inproceedings{WHURS19,
  title={Structural High-Resolution Satellite Image Indexing},
  author={Xia, Gui-Song and Yang, Wen and Delon, Julie and Gousseau, Yann and Sun, Hong and Ma{\^\i}tre, Henri},
  booktitle={ISPRS TC VII Symposium -- 100 Years {ISPRS}},
  pages={298--303},
  year={2010}
}

@inproceedings{LEEP,
  title={{LEEP}: A New Measure to Evaluate Transferability of Learned Representations},
  author={Nguyen, Cuong and Hassner, Tal and Seeger, Matthias and Archambeau, Cedric},
  booktitle={Proceedings of the 37th International Conference on Machine Learning ({ICML})},
  volume={119},
  pages={7294--7305},
  year={2020}
}

@inproceedings{HScore,
  title={An Information-Theoretic Approach to Transferability in Task Transfer Learning},
  author={Bao, Yajie and Li, Yang and Huang, Shao-Lun and Zhang, Lin and Zheng, Lizhong and Zamir, Amir R. and Guibas, Leonidas J.},
  booktitle={2019 IEEE International Conference on Image Processing ({ICIP})},
  pages={2309--2313},
  year={2019},
  doi={10.1109/ICIP.2019.8803726}
}

@inproceedings{LogME,
  title={{LogME}: Practical Assessment of Pre-trained Models for Transfer Learning},
  author={You, Kaichao and Liu, Yong and Wang, Jianmin and Long, Mingsheng},
  booktitle={Proceedings of the 38th International Conference on Machine Learning ({ICML})},
  volume={139},
  pages={12133--12143},
  year={2021}
}

@inproceedings{TransRate,
  title={Frustratingly Easy Transferability Estimation},
  author={Huang, Long-Kai and Huang, Junzhou and Rong, Yu and Yang, Qiang and Wei, Ying},
  booktitle={Proceedings of the 39th International Conference on Machine Learning ({ICML})},
  volume={162},
  pages={9201--9225},
  year={2022}
}

@inproceedings{Bolya2021PARC,
  title={Scalable Diverse Model Selection for Accessible Transfer Learning},
  author={Bolya, Daniel and Mittapalli, Rohit and Hoffman, Judy},
  booktitle={Advances in Neural Information Processing Systems ({NeurIPS})},
  volume={34},
  pages={19301--19312},
  year={2021}
}

@inproceedings{Agostinelli2022,
  title={Transferability Metrics for Selecting Source Model Ensembles},
  author={Agostinelli, Andrea and Uijlings, Jasper and Mensink, Thomas and Ferrari, Vittorio},
  booktitle={Proceedings of the IEEE/CVF Conference on Computer Vision and Pattern Recognition ({CVPR})},
  pages={7936--7946},
  year={2022}
}

@inproceedings{Agostinelli2022StableTM,
  title={How Stable Are Transferability Metrics Evaluations?},
  author={Agostinelli, Andrea and P{\'a}ndy, Michal and Uijlings, Jasper and Mensink, Thomas and Ferrari, Vittorio},
  booktitle={Proceedings of the European Conference on Computer Vision ({ECCV})},
  pages={303--321},
  year={2022}
}

@article{GroundedSAM,
  title={{Grounded SAM}: Assembling Open-World Models for Diverse Visual Tasks},
  author={Ren, Tianhe and Liu, Shilong and Zeng, Ailing and Lin, Jing and Li, Kunchang and Cao, He and Chen, Jiayu and Huang, Xinyu and Chen, Yukang and Yan, Feng and Zeng, Zhaoyang and Zhang, Hao and Li, Feng and Yang, Jie and Li, Hongyang and Jiang, Qing and Zhang, Lei},
  journal={arXiv preprint arXiv:2401.14159},
  year={2024}
}

@article{SAM3,
  title={{SAM} 3: Segment Anything with Concepts},
  author={Carion, Nicolas and Gustafson, Laura and Hu, Yuan-Ting and Debnath, Shoubhik and Hu, Ronghang and Suris, Didac and Ryali, Chaitanya and Alwala, Kalyan Vasudev and others},
  journal={arXiv preprint arXiv:2511.16719},
  year={2025}
}

@inproceedings{SEEM,
  title={Segment Everything Everywhere All at Once},
  author={Zou, Xueyan and Yang, Jianwei and Zhang, Hao and Li, Feng and Li, Linjie and Wang, Jianfeng and Wang, Lijuan and Gao, Jianfeng and Lee, Yong Jae},
  booktitle={Advances in Neural Information Processing Systems ({NeurIPS})},
  volume={36},
  year={2023}
}

@inproceedings{ODISE,
  title={Open-Vocabulary Panoptic Segmentation with Text-to-Image Diffusion Models},
  author={Xu, Jiarui and Liu, Sifei and Vahdat, Arash and Byeon, Wonmin and Wang, Xiaolong and De Mello, Shalini},
  booktitle={Proceedings of the IEEE/CVF Conference on Computer Vision and Pattern Recognition ({CVPR})},
  pages={2955--2966},
  year={2023}
}

@inproceedings{WinningTeam2023,
  title={Building a Winning Team: Selecting Source Model Ensembles using a Submodular Transferability Estimation Approach},
  author={B, Vimal K and Bachu, Saketh and Garg, Tanmay and Narasimhan, Niveditha Lakshmi and Konuru, Raghavan and Balasubramanian, Vineeth N},
  booktitle={Proceedings of the IEEE/CVF International Conference on Computer Vision ({ICCV})},
  pages={11609--11620},
  year={2023}
}

@article{li2026rethinking,
  title={Rethinking Model Selection in {VLM} Through the Lens of {Gromov-Wasserstein} Distance},
  author={Li, Muyang and Liu, Yucheng and Ma, Jianbo and Osborne, Elliot and Han, Bo and Liu, Tongliang},
  journal={arXiv preprint arXiv:2605.01325},
  year={2026}
}

@inproceedings{jia2021scaling,
  title={Scaling Up Visual and Vision-Language Representation Learning With Noisy Text Supervision},
  author={Jia, Chao and Yang, Yinfei and Xia, Ye and Chen, Yi-Ting and Parekh, Zarana and Pham, Hieu and Le, Quoc V. and Sung, Yun-Hsuan and Li, Zhen and Duerig, Tom},
  booktitle={Proceedings of the 38th International Conference on Machine Learning ({ICML})},
  series={Proceedings of Machine Learning Research},
  volume={139},
  pages={4904--4916},
  year={2021}
}

@inproceedings{mu2022slip,
  title={{SLIP}: Self-Supervision Meets Language-Image Pre-Training},
  author={Mu, Norman and Kirillov, Alexander and Wagner, David and Xie, Saining},
  booktitle={Proceedings of the European Conference on Computer Vision ({ECCV})},
  pages={529--544},
  year={2022}
}

@inproceedings{li2022supervision,
  title={Supervision Exists Everywhere: A Data Efficient Contrastive Language-Image Pre-training Paradigm},
  author={Li, Yangguang and Liang, Feng and Zhao, Lichen and Cui, Yufeng and Ouyang, Wanli and Shao, Jing and Yu, Fengwei and Yan, Junjie},
  booktitle={Proceedings of the International Conference on Learning Representations ({ICLR})},
  year={2022}
}

@inproceedings{singh2022flava,
  title={{FLAVA}: A Foundational Language and Vision Alignment Model},
  author={Singh, Amanpreet and Hu, Ronghang and Goswami, Vedanuj and Couairon, Guillaume and Galuba, Wojciech and Rohrbach, Marcus and Kiela, Douwe},
  booktitle={Proceedings of the IEEE/CVF Conference on Computer Vision and Pattern Recognition ({CVPR})},
  pages={15617--15629},
  year={2022}
}

@article{yu2022coca,
  title={{CoCa}: Contrastive Captioners Are Image-Text Foundation Models},
  author={Yu, Jiahui and Wang, Zirui and Vasudevan, Vijay and Yeung, Legg and Seyedhosseini, Mojtaba and Wu, Yonghui},
  journal={Transactions on Machine Learning Research},
  year={2022}
}

@article{koukounas2024jinaclipv2,
  title={{Jina-CLIP-v2}: Multilingual Multimodal Embeddings for Text and Images},
  author={Koukounas, Andreas and Mastrapas, Georgios and Eslami, Sedigheh and Wang, Bo and Akram, Mohammad Kalim and G{\"u}nther, Michael and Mohr, Isabelle and Sturua, Saba and Wang, Nan and Xiao, Han},
  journal={arXiv preprint arXiv:2412.08802},
  year={2024}
}

@article{bolya2025perception,
  title={Perception Encoder: The Best Visual Embeddings Are Not at the Output of the Network},
  author={Bolya, Daniel and Huang, Po-Yao and Sun, Peize and Cho, Jang Hyun and Madotto, Andrea and Wei, Chen and Ma, Tengyu and Zhi, Jiale and Rajasegaran, Jathushan and Rasheed, Hanoona and Wang, Junke and Monteiro, Marco and Xu, Hu and Dong, Shiyu and Ravi, Nikhila and Li, Daniel and Doll{\'a}r, Piotr and Feichtenhofer, Christoph},
  journal={arXiv preprint arXiv:2504.13181},
  year={2025}
}

@article{faghri2025mobileclip2,
  title={{MobileCLIP2}: Improving Multi-Modal Reinforced Training},
  author={Faghri, Fartash and Vasu, Pavan Kumar Anasosalu and Koc, Cem and Shankar, Vaishaal and Toshev, Alexander and Tuzel, Oncel and Pouransari, Hadi},
  journal={Transactions on Machine Learning Research},
  year={2025}
}

@inproceedings{silva2024multilingual,
  title={Multilingual Vision-Language Pre-training for the Remote Sensing Domain},
  author={Silva, Jo{\~a}o Daniel and Magalh{\~a}es, Jo{\~a}o and Tuia, Devis and Martins, Bruno},
  booktitle={Proceedings of the 32nd ACM International Conference on Advances in Geographic Information Systems ({SIGSPATIAL})},
  pages={220--232},
  year={2024},
  doi={10.1145/3678717.3691318}
}

@inproceedings{terlizzi2025rsdix,
  title={{RSDiX}: Lightweight and Data-Efficient {VLM}s for Remote Sensing Through Self-Distillation},
  author={Terlizzi, Andrea and Nazzaro, Angelo and Bernardi, Lorenzo and Bardozzo, Francesco and Tagliaferri, Roberto},
  booktitle={Proceedings of the International Joint Conference on Neural Networks ({IJCNN})},
  pages={1--10},
  year={2025},
  doi={10.1109/IJCNN64981.2025.11228553}
}

@article{haas2023learning,
  title={Learning Generalized Zero-Shot Learners for Open-Domain Image Geolocalization},
  author={Haas, Lukas and Alberti, Silas and Skreta, Michal},
  journal={arXiv preprint arXiv:2302.00275},
  year={2023}
}

@inproceedings{lin2023pmcclip,
  title={{PMC-CLIP}: Contrastive Language-Image Pre-training Using Biomedical Documents},
  author={Lin, Weixiong and Zhao, Ziheng and Zhang, Xiaoman and Wu, Chaoyi and Zhang, Ya and Wang, Yanfeng and Xie, Weidi},
  booktitle={Medical Image Computing and Computer Assisted Intervention -- MICCAI 2023},
  pages={525--536},
  year={2023}
}

@inproceedings{sun2025pathgen,
	title={{PathGen-1.6M}: 1.6 Million Pathology Image-Text Pairs Generation Through Multi-Agent Collaboration},
	author={Sun, Yuxuan and Zhang, Yunlong and Si, Yixuan and Zhu, Chenglu and Zhang, Kai and Shui, Zhongyi and Li, Jingxiong and Gong, Xuan and Lyu, Xinheng and Lin, Tao and Yang, Lin},
	booktitle={Proceedings of the International Conference on Learning Representations ({ICLR})},
	volume={2025},
	pages={94611--94653},
	year={2025}
}

\end{document}